%% file: main.tex
\documentclass[10pt,twocolumn,letterpaper]{article}
\usepackage{iccv}
\usepackage{times}
\usepackage{amsmath}
\usepackage{amssymb}
\usepackage{amsthm}
\usepackage{bm}
\usepackage{booktabs}
\usepackage{graphicx}
\usepackage{caption}
\usepackage{subcaption}
\usepackage{listings}

\usepackage[pagebackref=true,breaklinks=true,letterpaper=true,colorlinks,bookmarks=false]{hyperref}
\usepackage{cleveref}

\newtheorem{lemma}{Lemma}

\makeatletter
\renewcommand{\paragraph}{%
  \@startsection{paragraph}{4}%
  {\z@}{0.5em}{-1em}%
  {\normalfont\normalsize\bfseries}%
}
\makeatother

\crefname{section}{Sec.}{Sec.}
\Crefname{section}{Sec.}{Sec.}
\crefname{equation}{eq.}{eq.}
\Crefname{equation}{Eq.}{Eq.}

\newcommand{\bx}{{\bm x}}
\newcommand{\bbm}{{\bm m}}
\newcommand{\bbk}{{\bm k}}
\newcommand{\bbw}{{\bm w}}
\newcommand{\smax}{\operatornamewithlimits{smax}}

\newcommand{\argmax}{\operatornamewithlimits{argmax}}
\newcommand{\sortvec}{\operatorname{vecsort}}
\renewcommand{\b}[1]{\textbf{#1}}
\renewcommand{\u}[1]{\underline{#1}}

\newcommand\blfootnote[1]{%
  \begingroup
  \renewcommand\thefootnote{}\footnote{#1}%
  \addtocounter{footnote}{-1}%
  \endgroup
}

\iccvfinalcopy
\def\iccvPaperID{3949} 
\ificcvfinal\fi

\title{Understanding Deep Networks via Extremal Perturbations and Smooth Masks}
\author{
  Ruth Fong$^\dagger$\thanks{Work done as a contractor at FAIR. $^\dagger$ denotes equal contributions.}\\
  University of Oxford
  \and
  Mandela Patrick$^\dagger$\\
  University of Oxford
  \and
  Andrea Vedaldi\\
  Facebook AI Research
}

\begin{document}
\maketitle
\begin{abstract}
The problem of attribution is concerned with identifying the parts of an input that are responsible for a model's output.
An important family of attribution methods is based on measuring the effect of perturbations applied to the input.
In this paper, we discuss some of the shortcomings of existing approaches to perturbation analysis and address them by introducing the concept of extremal perturbations, which are theoretically grounded and interpretable.
We also introduce a number of technical innovations to compute extremal perturbations, including a new area constraint and a parametric family of smooth perturbations, which allow us to remove all tunable hyper-parameters from the optimization problem.
We analyze the effect of perturbations as a function of their area, demonstrating excellent sensitivity to the spatial properties of the deep neural network under stimulation.
We also extend perturbation analysis to the intermediate layers of a  network.
This application allows us to identify the salient channels necessary for classification, which, when visualized using feature inversion, can be used to elucidate model behavior.
Lastly, we introduce TorchRay\footnote{\url{github.com/facebookresearch/TorchRay}}, an interpretability library built on PyTorch.
\end{abstract}
\input{intro}

\input{related}
\input{method}

\input{experiments}

\input{intermediate}

\input{conclusions}
\blfootnote{\noindent\textbf{Acknowledgements.}
We are grateful for support from the Open Philanthropy Project (R.F.), the Rhodes Trust (M.P.), and ESPRC EP/L015897/1 (CDT in Autonomous Intelligent Machines and Systems) (M.P).
We also thank Jianming Zhang and Samuel Albanie for help on re-implementing the Pointing Game~\cite{zhang2016excitation} in PyTorch.}
{\small\bibliographystyle{ieee_fullname}\bibliography{refs}}
\input{appendix}
\end{document}

%% file: intro.tex
\section{Introduction}\label{s:intro}

Deep networks often have excellent prediction accuracy, but the basis of their predictions is usually difficult to understand.
\emph{Attribution} aims at characterising the response of neural networks by finding which parts of the network's input are the most responsible for determining its output.
Most attribution methods are based on backtracking the network's activations from the output back to the input, usually via a modification of the backpropagation algorithm~\cite{simonyan14deep, zeiler2014visualizing, springenberg2014striving, zhang2016excitation, selvaraju17gradcam, bach2015pixel}.
When applied to computer vision models, these methods result in \emph{saliency maps} that highlight important regions in the input image.

However, most attribution methods do not start from a definition of what makes an input region important for the neural network.
Instead, most saliency maps are validated \emph{a-posteriori} by either showing that they correlate with the image content (e.g., by highlighting relevant object categories), or that they find image regions that, if perturbed, have a large effect on the network's output (see~\cref{s:related}).

\input{fig-splash}
\input{fig-comparisons}

Some attribution methods, on the other hand, directly perform an analysis of the effect of \emph{perturbing} the network's input on its output~\cite{zeiler2014visualizing,Petsiuk2018rise,fong17interpretable,dabkowski2017real}.
This usually amounts to selectively deleting (or preserving)  parts of the input and observing the effect of that change to the model's output.
The advantage is that the meaning of such an analysis is clear from the outset.
However, this is not as straightforward as it may seem on a first glance.
First, since it is not possible to visualise \emph{all} possible perturbations, one must find \emph{representative} ones.
Since larger perturbations will have, on average, a larger effect on the network, one is usually interested in small perturbations with a large effect (or large perturbations with a small effect).
Second, Fong and Vedaldi~\cite{fong17interpretable} show that searching for perturbations with a large effect on the network's output usually results in \emph{pathological} perturbations that trigger adversarial effects in the network.
Characterizing instead the \emph{typical} behavior of the model requires restricting the search to more representative perturbations via regularization terms.
This results in an optimization problem that trades off maximizing the effect of the perturbation with its smoothness and size.
In practice, this trade off is difficult to control numerically and somewhat obscures the meaning of the analysis.

In this paper, we make three contributions.
First, instead of mixing several effects in a single energy term to optimize as in Fong and Vedaldi~\cite{fong17interpretable}, we introduce the concept of \emph{extremal perturbations}.
A perturbation is extremal if it has maximal effect on the network's output among all perturbations of a given, fixed area.
Furthermore, the perturbations are regularised by choosing them within family with a minimum guaranteed level of smoothness.
In this way, the optimisation is carried over the perturbation effect only, without having to balance several energy terms as done in~\cite{fong17interpretable}.
Lastly, by sweeping the area parameter, we can study the perturbation's effect w.r.t. its size.

The second contribution is technical and is to provide a concrete algorithm to calculate the extremal perturbations.
First, in the optimisation we must \emph{constrain} the perturbation size to be equal to a target value.
To this end, we introduce a new ranking-based \emph{area loss} that can enforce these type of constraints in a stable and efficient manner.
This loss, which we believe can be beneficial beyond our perturbation analysis, can be interpreted as a hard constraint, similar to a logarithmic barrier, differing from the soft penalty on the area in Fong and Vedaldi~\cite{fong17interpretable}.
Second, we construct a parametric family of perturbations with a minimum guarantee amount of smoothness.
For this, we use the \emph{(smooth)-max-convolution operator} and a \emph{perturbation pyramid}.

As a final contribution, we extend the framework of perturbation analysis to the intermediate activations of a deep neural network rather than its input.
This allows us to explore how perturbations can be used beyond spatial, input-level attribution, to channel, intermediate-layer attribution. When combined with existing visualization techniques such as feature inversion~\cite{mahendran15understanding,olah2017feature,mordvintsev2018differentiable,ulyanov2018deep}, we demonstrate how intermediate-layer perturbations can help us understand which channels are salient for classification.

%% file: fig-splash.tex
\begin{figure}\centering
\includegraphics[width=\linewidth]{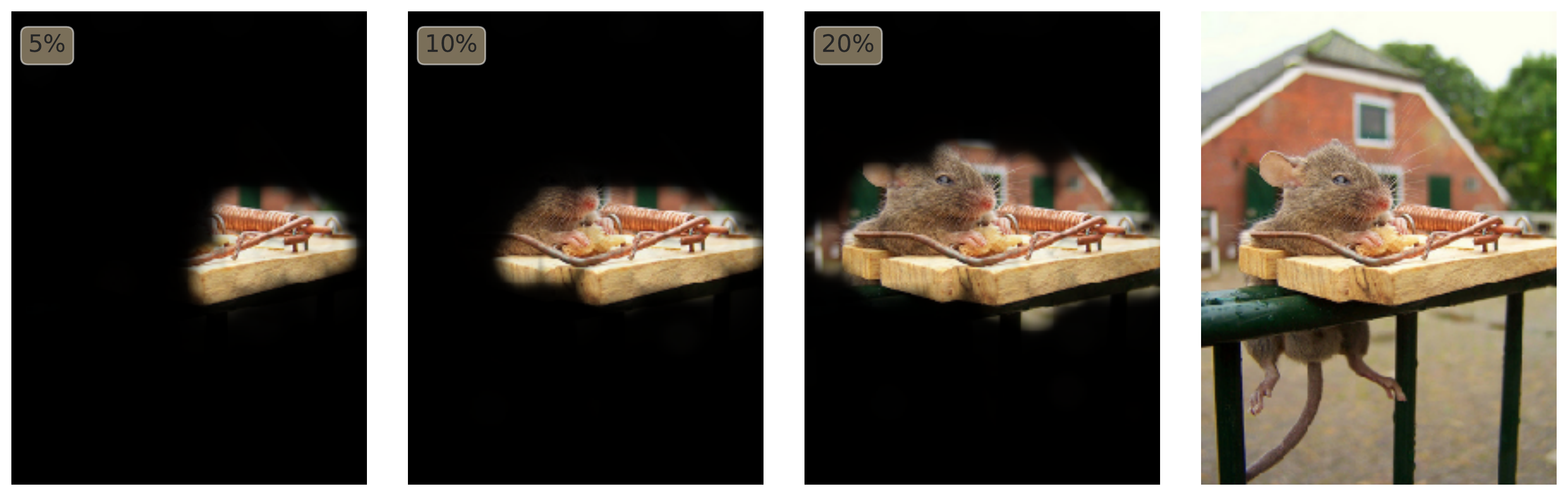}
\caption{\textbf{Extremal perturbations} are regions of an image that, for a given area (boxed), maximally affect the activation of a certain neuron in a neural network (i.e., ``mousetrap'' class score).
As the area of the perturbation is increased, the method reveals more of the image, in order of decreasing importance.
For clarity, we black out the masked regions; in practice, the network sees blurred regions.
}\label{f:splash}
\end{figure}

%% file: fig-comparisons.tex
\begin{figure*}[ht]
\centering
\includegraphics[width=0.99\linewidth]{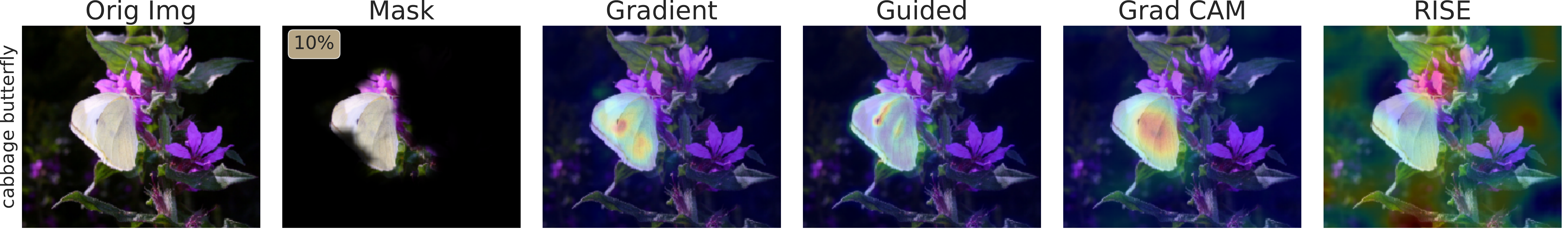}
\includegraphics[width=0.99\linewidth]{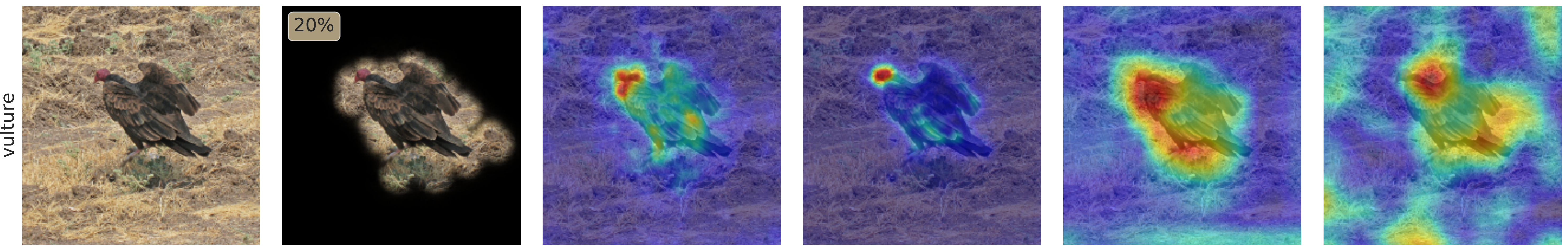}
\includegraphics[width=0.99\linewidth]{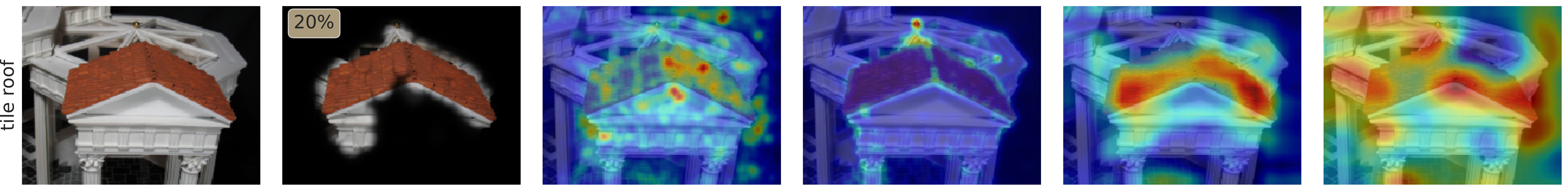}
\includegraphics[width=0.99\linewidth]{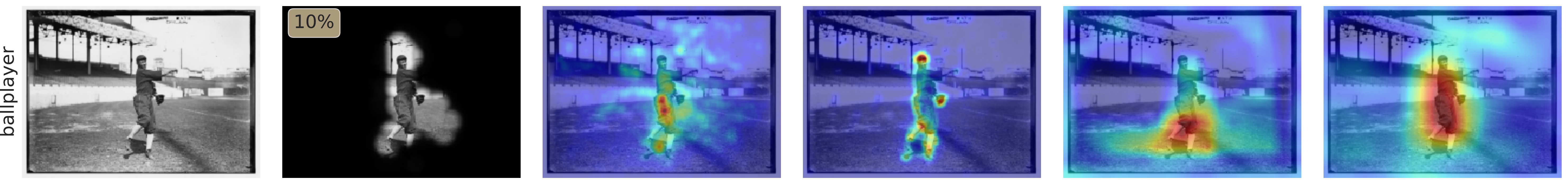}
\includegraphics[width=0.99\linewidth]{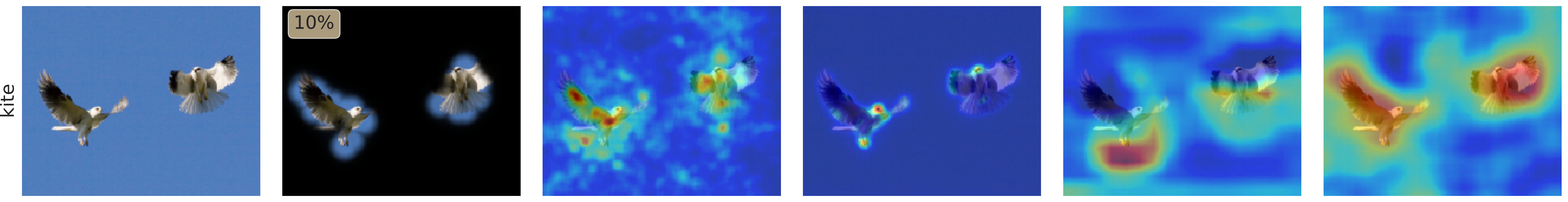}
\caption{\textbf{Comparison with other attribution methods.} We compare our extremal perturbations (optimal area $a^*$ in box) to several popular attribution methods: gradient~\cite{simonyan14deep}, guided backpropagation~\cite{springenberg2014striving}, Grad-CAM~\cite{selvaraju17gradcam}, and RISE~\cite{Petsiuk2018rise}.
}\label{fig:comparison}
\end{figure*}

%% file: related.tex
\section{Related work}\label{s:related}

\paragraph{Backpropagation-based methods.}

Many attribution techniques leverage backpropagation to track information from the network's output back to its input, or an intermediate layer.
Since they are based on simple modifications of the backpropagation algorithm, they only require a single forward and backward pass through the model, and are thus efficient.
\cite{simonyan14deep}'s gradient method, which uses unmodified backprop, visualizes the derivative of the network's output w.r.t.~the input image.
Other works (e.g., DeCovNet~\cite{zeiler2014visualizing}, Guided Backprop~\cite{springenberg2014striving}, and SmoothGrad~\cite{smilkov2017smoothgrad}) reduce the noise in the gradient signal by tweaking the backprop rules of certain layers.
Other methods generate visualizations by either combining gradients, network weights and/or activations at a specific layer (e.g., CAM~\cite{zhou2016learning} and Grad-CAM~\cite{selvaraju17gradcam}) or further modify the backpropn rules to have a probabilistic or local approximation interpretation (e.g., LRP~\cite{bach2015pixel} and Excitation Backprop~\cite{zhang2016excitation}).

Several papers have shown that some (but not all) backpropagation-based methods produce the same saliency map regardless of the output neuron being analysed~\cite{mahendran16salient}, or even regardless of network parameters~\cite{Adebayo2018Sanity}.
Thus, such methods may capture average network properties but may not be able to characterise individual outputs or intermediate activations, or in some cases the model parameters.

\paragraph{Perturbation-based methods.}

Another family of approaches perturbs the inputs to a model and observes resultant changes to the outputs.
Occlusion~\cite{zeiler2014visualizing} and RISE~\cite{Petsiuk2018rise} occlude an image using regular or random occlusions patterns, respectively, and weigh the changes in the output by the occluding patterns.
Meaningful perturbations~\cite{fong17interpretable} optimize a spatial perturbation mask that maximally affects a model's output.
Real-time saliency~\cite{dabkowski2017real} builds on~\cite{fong17interpretable} and learns to predict such a perturbation mask with a second neural network.
Other works have leveraged perturbations at the input~\cite{singh2017hide, wei2017object} and intermediate layers~\cite{wang2017fast} to perform weakly and fully supervised localization.

\paragraph{Approximation-based methods.}

Black-box models can be analyzed by approximating them (locally) with simpler, more interpretable models.
The gradient method of~\cite{simonyan14deep} and, more explicitly, LIME~\cite{ribeiro2016should}, do so using linear models.
Approximations using decision trees or other models are also possible, although less applicable to visual inputs.

\paragraph{Visualizations of intermediate activations.}

To characterize a filter's behavior, Zeiler and Fergus~\cite{zeiler2014visualizing} show dataset examples from the training set that maximally activate that filter.
Similarly, activation maximization~\cite{simonyan14deep} learns an input image that maximally activates a filter.
Feature inversion~\cite{mahendran15understanding} learns an image that reconstructs a network's intermediate activations while leveraging a natural image prior for visual clarity.
Subsequent works tackled the problem of improving the natural image prior for feature inversion and/or activation maximization~\cite{ulyanov2018deep,olah2017feature,mordvintsev2018differentiable,nguyen2016synthesizing,nguyen2017plug}.
Recently, some methods have measured the performance of single~\cite{bau17network,zhou2018revisiting} and combinations of~\cite{kim2017interpretability,fong18net2vec} filter activations on probe tasks like classification and segmentation to identify which filter(s) encode what concepts.

One difficulty in undertaking channel attribution is that, unlike spatial attribution, where a salient image region is naturally interpretable to humans, simply identifying ``important channels'' is insufficient as they are not naturally interpretable. To address this, we combine the aforementioned visualization techniques with channel attribution.

%% file: method.tex
\section{Method}\label{s:method}

We first summarize the perturbation analysis of~\cite{fong17interpretable} and then introduce our extremal perturbations framework.

\input{fig-area-growing}

\subsection{Perturbation analysis}\label{s:analysis}

Let $\bx :\Omega\rightarrow\mathbb{R}^3$ be a colour image, where $\Omega=\{0,\dots,H-1\}\times\{0,\dots,W-1\}$ is a discrete lattice, and let $\Phi$ be a model, such as a convolutional neural network, that maps the image to a scalar output value $\Phi(\bx) \in \mathbb{R}$.
The latter could be an output activation, corresponding to a class prediction score, in a model trained for image classification, or an intermediate activation.

In the following, we investigate which parts of the input $\bx$ strongly excite the model, causing the response $\Phi(\bx)$ to be large.
In particular, we would like to find a \emph{mask} $\bbm$ assigning to each pixel $u\in\Omega$ a value $\bbm(u) \in \{0,1\}$, where $\bbm(u)=1$ means that the pixel strongly contributes to the output and $\bbm(u)=0$ that it does not.

In order to assess the importance of a pixel, we use the mask to induce a local perturbation of the image, denoted $\hat\bx = \bbm \otimes \bx$.
The details of the perturbation model are discussed below, but for now it suffices to say that pixels for which $\bbm(u)=1$ are preserved, whereas the others are blurred away.
The goal is then to find a small subset of pixels that, when preserved, are sufficient to retain a large value of the output $\Phi(\bbm\otimes\bx)$.

Fong and Vedaldi~\cite{fong17interpretable} propose to identify such salient pixels by solving an optimization problem of the type:
\begin{equation}\label{e:orig}
\bbm_{\lambda,\beta} = \argmax_{\bbm} \Phi(\bbm \otimes \bx) - \lambda \|\bbm\|_1 - \beta \mathcal{S}(\bbm).
\end{equation}
The first term encourages the network's response to be large.
The second encourages the mask to select a small part of the input image, blurring as many pixels as possible.
The third further regularises the smoothness of the mask by penalising irregular shapes.

The problem with this formulation is that the meaning of the trade-off established by optimizing~\cref{e:orig} is unclear as the three terms, model response, mask area and mask regularity, are not commensurate.
In particular, choosing different $\lambda$ and $\beta$ values in~\cref{e:orig} will result in different masks without a clear way of comparing them.

\subsection{Extremal perturbations}\label{s:extremal}

In order to remove the balancing issues with~\cref{e:orig}, we propose to constrain the area of the mask to a fixed value (as a fraction $a|\Omega|$ of the input image area).
Furthermore, we control the smoothness of the mask by choosing it in a fixed set $\mathcal{M}$ of sufficiently smooth functions.
Then,  we find the mask of that size that maximizes the model's output:
\begin{equation}\label{e:areac}
\bbm_a = \argmax_{\bbm:~\|\bbm\|_1=a|\Omega|,~\bbm \in \mathcal{M}} \Phi(\bbm \otimes \bx).
\end{equation}
Note that the resulting mask is a function of the chosen area $a$ only.
With this, we can define the concept of \emph{extremal perturbation} as follows.
Consider a lower bound $\Phi_0$ on the model's output (for example we may set $\Phi_0 =\tau\Phi(\bx)$ to be a fraction $\tau$ of the model's output on the unperturbed image).
Then, we search for the \emph{smallest mask} that achieves at least this output level.
This amounts to sweeping the area parameter $a$ in~\cref{e:areac} to find
\begin{equation}\label{e:areamax}
a^* = \min\{ a : \Phi(\bbm_a \otimes \bx) \geq \Phi_0 \}.
\end{equation}
The mask $\bbm_{a^*}$ is extremal because preserving a smaller portion of the input image is not sufficient to excite the network's response above $\Phi_0$.
This is illustrated in~\cref{fig:area_growing}.

\paragraph{Interpretation.}

An extremal perturbation is a single mask $\bbm_{a^*}$ that results in a large model response, in the sense that $\Phi(\bbm_{a^*}\otimes\bx) \geq \Phi_0$.\
However, due to extremality, we \emph{also} know that any smaller mask does not result in an equally large response:
$
 \forall \bbm :~ \|\bbm\|_1 <  \|\bbm_{a^{*}}\|_1~\Rightarrow~ \Phi(\bbm \otimes \bx) < \Phi_0.
$
Hence, a single extremal mask is informative because it characterises a \emph{whole family} of input perturbations.

This connects extremal perturbations to methods like~\cite{ribeiro2016should,fong17interpretable}, which explain a network by finding a succinct and interpretable description of its input-output mapping.
For example, the gradient~\cite{simonyan14deep} and LIME~\cite{ribeiro2016should}  approximate the network locally around an input $\bx$ using the Taylor expansion $\Phi(\bx') \approx \langle \nabla \Phi(\bx), \bx' - \bx \rangle + \Phi(\bx)$; their explanation  is the gradient $\nabla \Phi(\bx)$ and their perturbations span a neighbourhood of $\bx$.

\paragraph{Preservation vs deletion.}\label{s:variants}

Formulation~\eqref{e:areac} is analogous to what~\cite{fong17interpretable} calls the ``preservation game'' as the goal is to find a mask that preserves (maximises) the model's response.
We also consider their ``deletion game'' obtaining by optimising
$
\Phi((1-\bbm) \otimes \bx)
$
in~\cref{e:areac}, so that the goal is to suppress the response when looking outside the mask, and the hybrid~\cite{dabkowski2017real}, obtained by optimising
$
\Phi(\bbm \otimes \bx) - \Phi((1-\bbm) \otimes \bx),
$ where the goal is to simultaneously preserve the response inside the mask and suppress it outside

\subsection{Area constraint}\label{s:area}

Enforcing the area constraint in~\cref{e:areac} is non-trivial; here, we present an effective approach to do so (other approaches like~\cite{kervadec2019constrained} do not encourage binary masks).
First, since we would like to optimize~\cref{e:areac} using a gradient-based method, we relax the mask to span the full range $[0,1]$.
Then, a possible approach would be to count how many values $\bbm(u)$ are sufficiently close to the value 1 and penalize masks for which this count deviates from the target value $a|\Omega|$.
However, this approach requires soft-counting, with a corresponding tunable parameter for binning.

In order to avoid such difficulties, we propose instead to \emph{vectorize and sort} in non-decreasing order the values of the mask $\bbm$, resulting in a vector $\sortvec(\bbm) \in [0,1]^{|\Omega|}$.
If the mask $\bbm$ satisfies the area constraint exactly, then the output of $\sortvec(\bbm)$ is a vector $\mathbf{r}_a \in [0,1]^{|\Omega|}$ consisting of $(1-a)|\Omega|$ zeros followed by $a|\Omega|$ ones.
This is captured by the regularization term:
$
R_a(\bbm) = \| \sortvec(\bbm) - \mathbf{r}_a \|^2.
$
We can then rewrite~\cref{e:areac} as
\begin{equation}\label{e:areac-relaxed}
\bbm_a = \argmax_{\bbm \in\mathcal{M}} \Phi(\bbm \otimes \bx) - \lambda R_a(\bbm).
\end{equation}
Note that we have reintroduced a weighting factor $\lambda$ in the formulation, so on a glance we have lost the advantage of formulation~\eqref{e:areac} over the one of~\cref{e:orig}.
In fact, this is not the case: during optimization we simply set $\lambda$ to be as large as numerics allow it as we expect the area constraint to be (nearly) exactly satisfied; similarly to a logarithmic barrier, $\lambda$ then has little effect on which mask $\bbm_a$ is found.%

\subsection{Perturbation operator}\label{s:perturb}

In this section we define the perturbation operator $\bbm \otimes \bx$.
To do so, consider a \emph{local perturbation operator}
$
\pi(\bx;u,\sigma) \in \mathbb{R}^3
$
that applies a perturbation of intensity $\sigma\geq 0$ to pixel $u\in\Omega$.
We assume that the lowest intensity $\sigma=0$ corresponds to no perturbation, i.e.~$\pi(\bx;u,0) = \bx(u)$.
Here we use as perturbations the Gaussian blur%
\footnote{
  Another choice is the fade-to-black perturbation which, for $0\leq \sigma \leq 1$, is given by
  $
  \pi_f(\bx;u,\sigma)
  =
  (1 - \sigma) \cdot \bx(u).
  $
  }
  $$
  \pi_g(\bx;u,\sigma)
  =
  \frac{
  \sum_{v \in \Omega} g_\sigma(u - v) \bx(v)
  }{
  \sum_{v \in \Omega} g_\sigma(u - v)
},
~~~
g_\sigma(u)
=
e^{-\frac{\|u\|^2}{2\sigma^2}}.
$$
The mask $\bbm$ then doses the perturbation spatially:
$
(\bbm \otimes \bx)(u) = \pi(\bx;u, \sigma_\text{max}\cdot (1-\bbm(u)))
$
where $\sigma_\text{max}$ is the maximum perturbation intensity.%
\footnote{
  For efficiency, this is implemented by generating a \emph{perturbation pyramid}
  $
  \pi(\bx; \cdot, \sigma_\text{max}\cdot l/L), ~l=0,\dots,L
  $
  that contains $L+1$ progressively more perturbed versions of the image.
  Then $\bbm \otimes \bx$ can be computed via bilinear interpolation by using $(u, \bbm(u))$ as an indices in the pyramid.
}

\subsection{Smooth masks}\label{s:maskparam}

Next, we define the space of smooth masks $\mathcal{M}$.
For this, we consider an auxiliary mask $\bar\bbm : \Omega \rightarrow [0,1]$.
Given that the range of $\bar\bbm$ is bounded, we can obtain a smooth mask $\bbm$ by convolving $\bar\bbm$ by a  Gaussian or similar kernel $\bbk : \Omega \rightarrow \mathbb{R}_+$%
\footnote{It is easy to show that in this case the derivative of the smoothed mask $\||\nabla (\bbk \ast \bar\bbm)\|| \leq \|\nabla \bbk \|$ is always less than the one of the kernel.} via the typical \emph{convolution operator}:
\begin{equation}\label{e:guassian}
\bbm(u)
=
Z^{-1}\sum_{v\in\Omega}
\bbk(u-v)\bar\bbm(v)
\end{equation}
where $Z$ normalizes the kernel to sum to one.
However, this has the issue that setting $\bar \bbm(u) =1$ does not necessarily result in $\bbm(u) = 1$ after filtering, and we would like our final mask to be (close to) binary.

To address this issue, we consider the \emph{max-convolution operator}:
\begin{equation}\label{e:max}
\bbm(u)
=
\max_{v\in\Omega}
\bbk(u-v)\bar\bbm(v).
\end{equation}
This solves the issue above while at the same time guaranteeing that the smoothed mask does not change faster than the smoothing kernel, as shown in the following lemma (proof in supp.~mat.).

\begin{lemma}
Consider functions $\bar\bbm, \bbk : \Omega \rightarrow [0,1]$ and let $\bbm$ be defined as in~\cref{e:max}.
If $\bbk(0) = 1$, then $\bar\bbm(u) \leq \bbm(u) \leq 1$ for all $u\in\Omega$; in particular, if $\bar\bbm(u)=1$, then $\bbm(u)=1$ as well.
Furthermore, if $\bbk$ is Lipschitz continuous with constant $K$,
then $\bbm$ is also Lipschitz continuous with a constant at most as large as $K$.
\end{lemma}

The $\max$ operator in~\cref{e:max} yields sparse gradients.
Thus, to facilitate optimization, we introduce the \emph{smooth max} operator\footnote{Not to be confused with the softmax with temperature, as in~\cite{hinton2015distilling}.}, $\smax$, to replace the $\max$ operator.
For a function $f(u),u\in\Omega$ and temperature $T > 0$:
\begin{equation}\label{e:smax_operator}
\smax_{u\in\Omega;T} f(u)
=
\frac
{\sum_{u\in\Omega} f(u)\exp{{f(u)}/{T}}}
{\sum_{u\in\Omega}\exp{{f(u)}/{T}}}
\end{equation}
The $\smax$ operator smoothly varies from behaving like the mean operator in~\cref{e:guassian} as $T \to \infty$ to behaving like the $\max$ operator as $T \to 0$ (see~\cref{fig:maxconv_1d}).
This operator is used instead of $\operatorname{max}$ in~\cref{e:max}.

\input{fig-maxconv-1d}

\paragraph{Implementation details.}
In practice, we use a smaller parameterization mask $\bar\bbm$ defined on a lattice $\bar \Omega = \{0,\dots,\bar H-1\} \times \{0,\dots,\bar W-1\}$, where the full-resolution mask $\bbm$ has dimensions $H = \rho \bar H$ and $W = \rho \bar W$.
We then modify~\eqref{e:max} to perform upsampling in the same way as the standard convolution transpose operator.

\input{fig-several-area-growths}

%% file: fig-area-growing.tex
\begin{figure*}[ht]\centering
\includegraphics[width=0.78\linewidth]{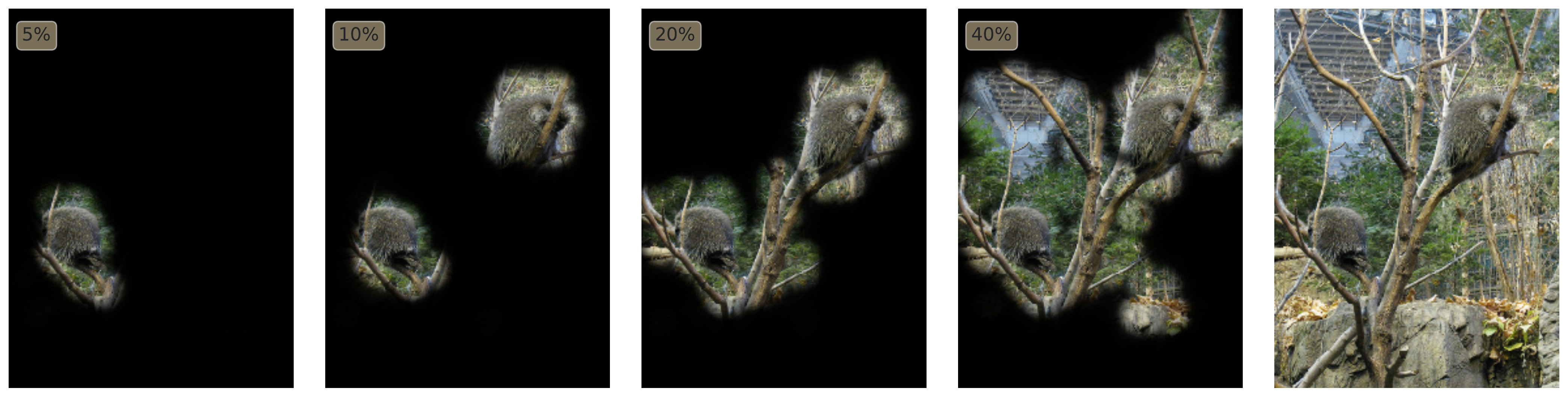}
\includegraphics[width=0.195\linewidth]{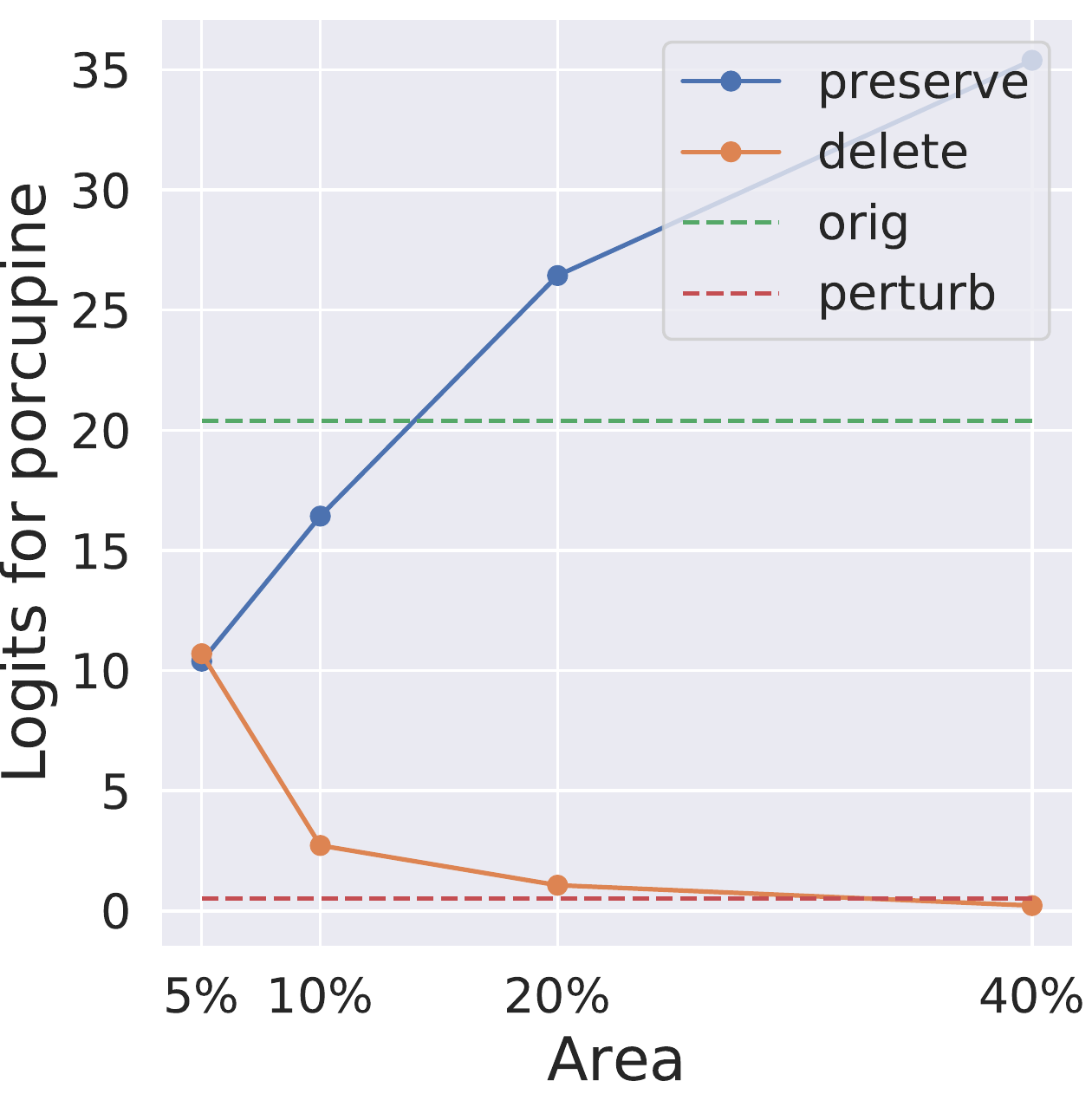}
\caption{\textbf{Extremal perturbations and monotonic effects.}
Left: ``porcupine'' masks computed for several areas $a$ ($a$ in box).
Right: $\Phi(\bbm_a \otimes \bx)$ (preservation; blue) and $\Phi((1-\bbm_a)\otimes\bx)$ (deletion; orange) plotted as a function of $a$.
At $a\approx 15\%$ the preserved region scores \emph{higher} than preserving the entire image (green).
At $a\approx 20\%$, perturbing the complementary region scores \emph{similarly} to fully perturbing the entire image (red).
}\label{fig:area_growing}
\end{figure*}

%% file: fig-maxconv-1d.tex
\begin{figure}
\centering
\includegraphics[width=0.9\linewidth]{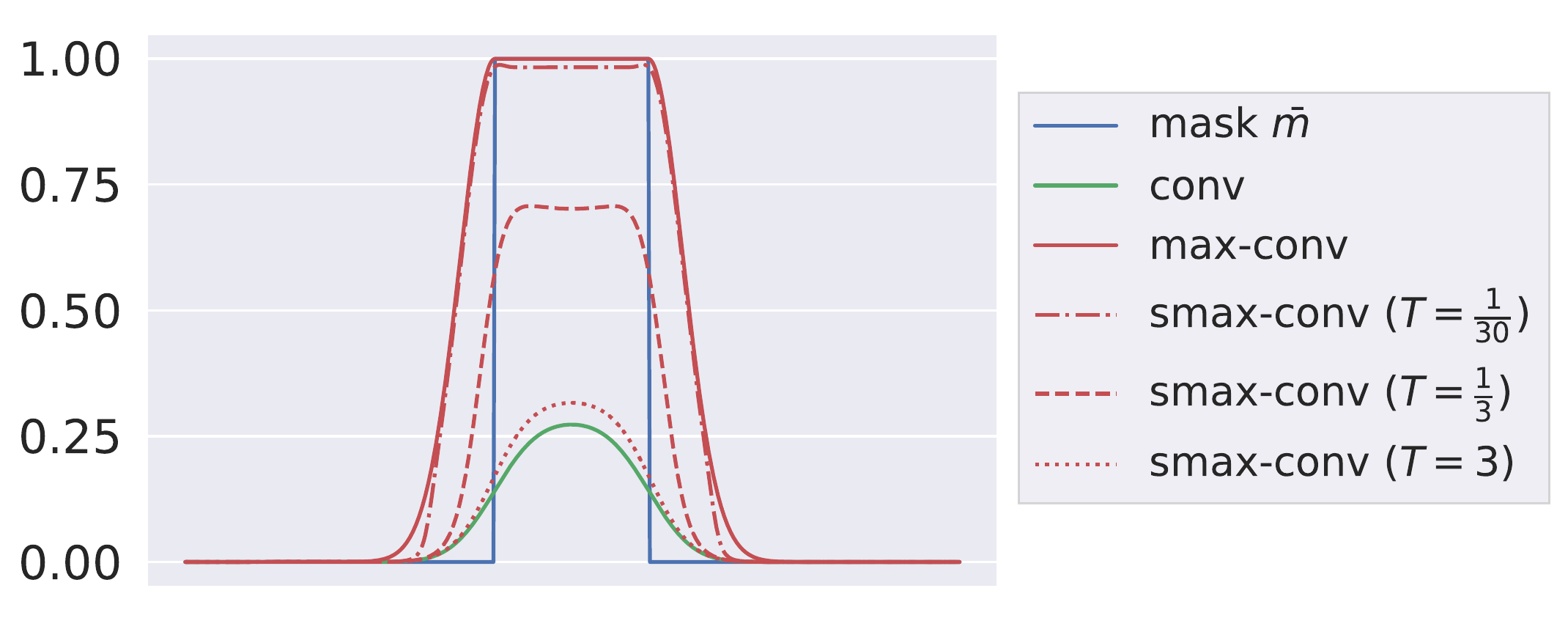}
\caption{\textbf{Convolution operators for smooth masks.}
Gaussian smoothing a mask (blue) with the typical convolution operator yields a dampened, smooth mask (green).
Our max-convolution operator mitigates this effect while still smoothing (red solid).
Our $\smax$ operator, which yields more distributed gradients than $\max$, varies between the other two convolution operators (red dotted).
}
\label{fig:maxconv_1d}
\end{figure}

%% file: fig-several-area-growths.tex
\begin{figure}
	\centering
	\includegraphics[width=\linewidth]{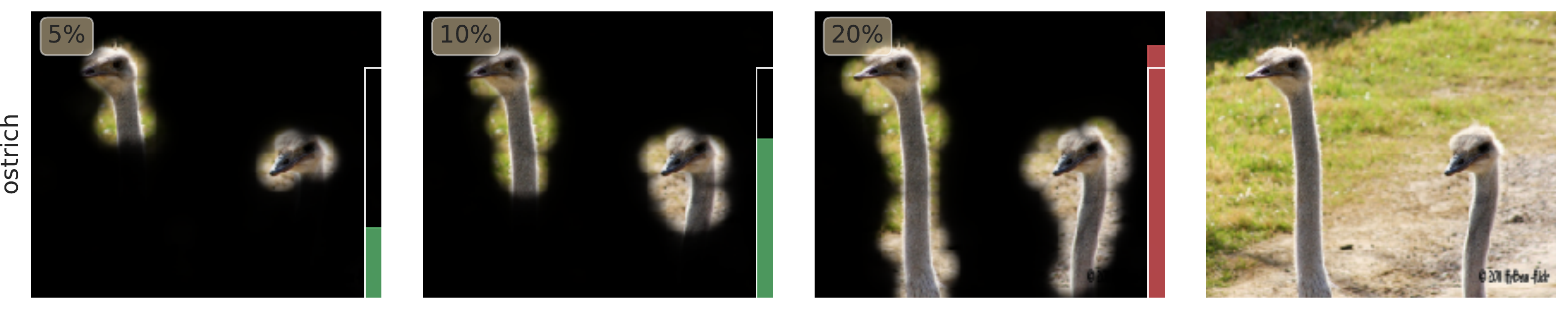}
	\includegraphics[width=\linewidth]{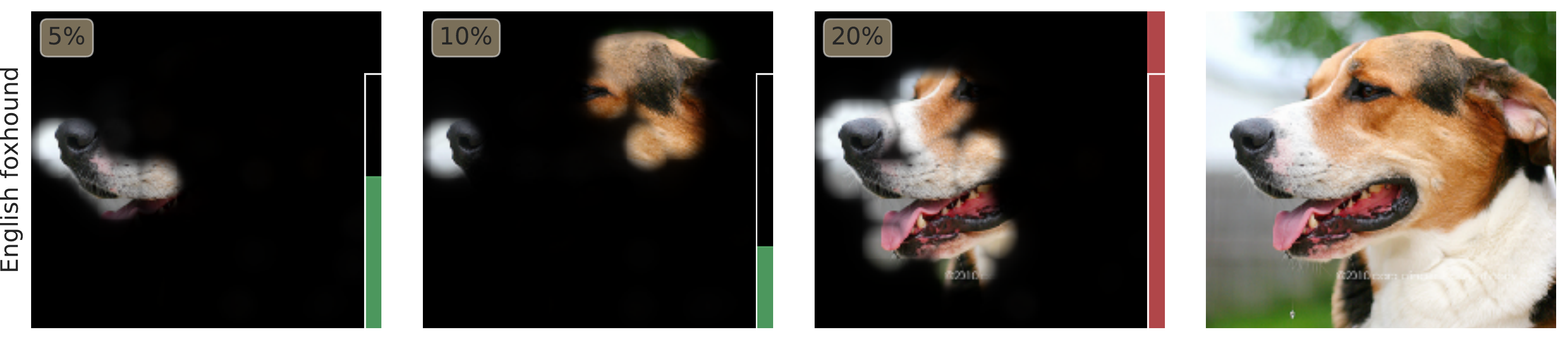}
	\includegraphics[width=\linewidth]{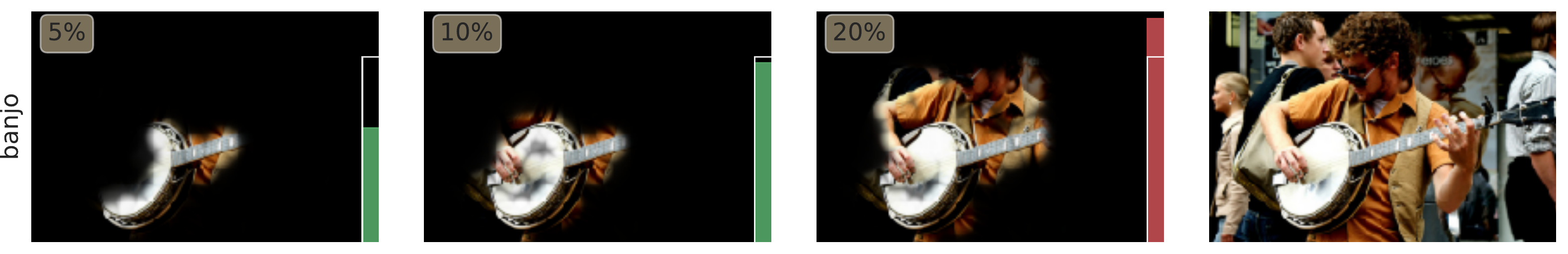}
	\includegraphics[width=\linewidth]{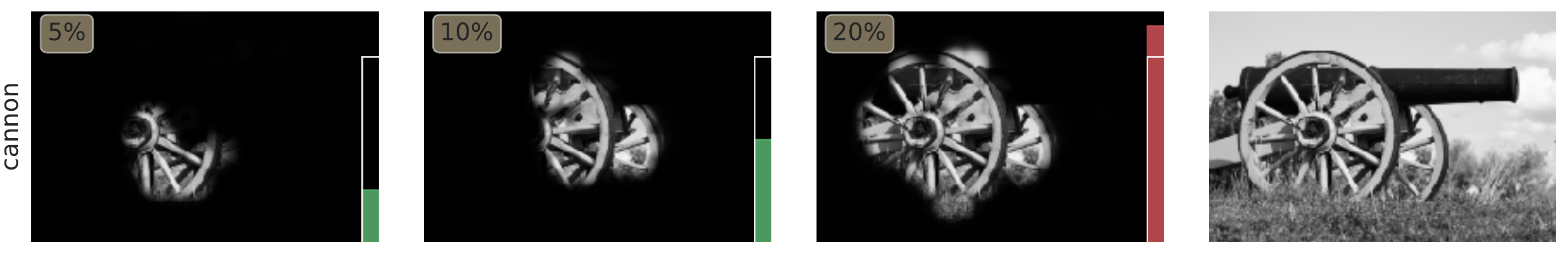}
	\includegraphics[width=\linewidth]{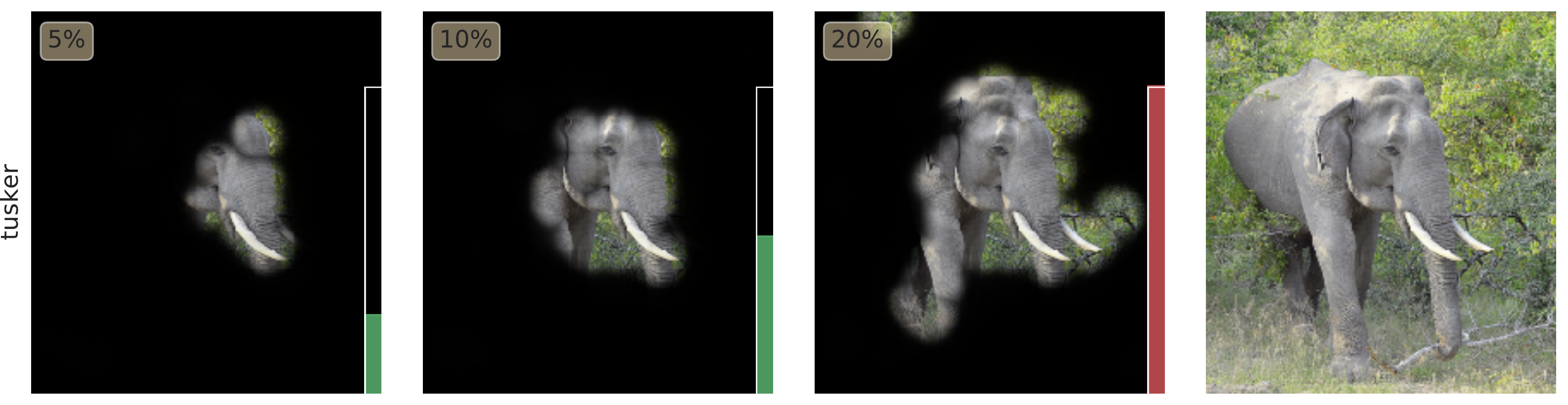}
	\includegraphics[width=\linewidth]{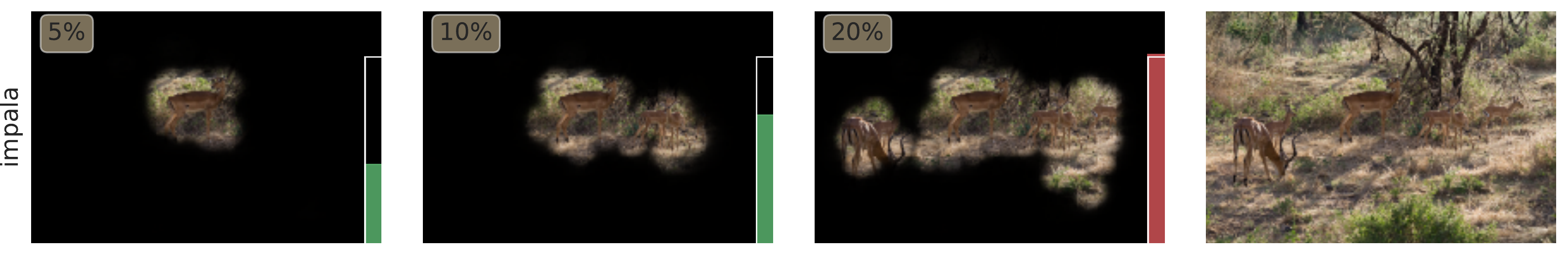}
	\includegraphics[width=\linewidth]{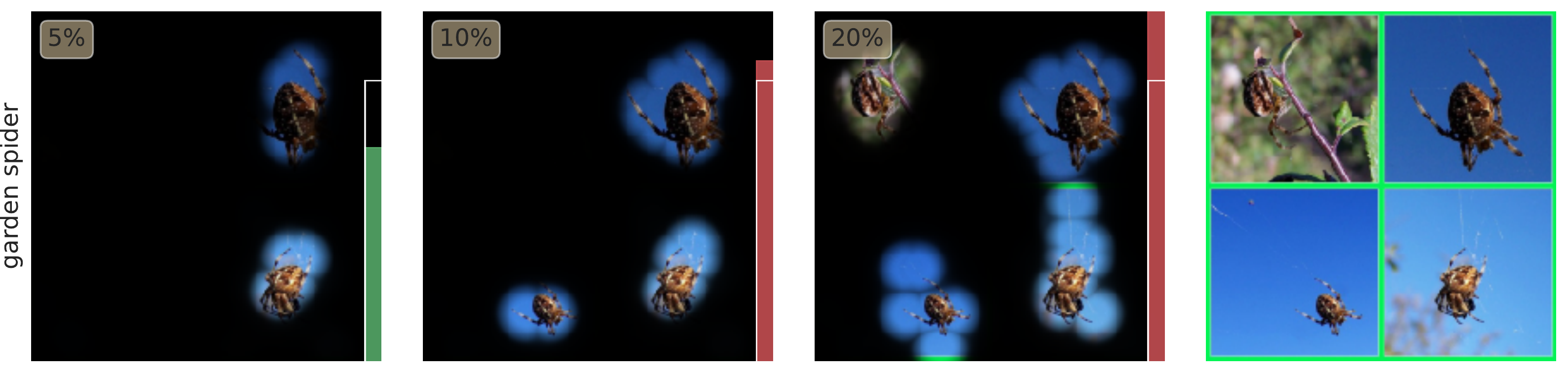}
	\caption{\textbf{Area growth.} Although each mask is learned independently, these plots highlight what the network considers to be most discriminative and complete. The bar graph visualizes $\Phi(\bm m_{a} \odot \bx)$ as a normalized fraction of $\Phi_0 = \Phi(\bx)$ (and saturates after exceeding $\Phi_0$ by $25\%$).}
	\label{fig:area_progressions}
\end{figure}

%% file: experiments.tex
\section{Experiments}\label{s:experiments}

\paragraph{Implementation details.}

Unless otherwise noted, all visualizations use the ImageNet validation set, the VGG16 network and the preservation formulation (\cref{s:extremal}).
Specifically, $\Phi(\bx)$ is the classification score (before softmax) that network associates to the ground-truth class in the image.
Masks are computed for areas $a \in\{0.05, 0.1, 0.2, 0.4, 0.6, 0.8\}$.
To determine the optimal area $a^*$ of the extremal perturbations~\eqref{e:areamax}, we set the threshold $\Phi_0 = \Phi(\bx)$ (which is the score on the unperturbed image).

Masks are optimised using SGD, initializing them with all ones (everything preserved). SGD uses momentum 0.9 and 1600 iterations.
$\lambda$ is set to 300 and doubled at 1/3 and 2/3 of the iterations and, in~\cref{e:smax_operator}, $1/T\approx 20$.
Before upsampling, the kernel $\bbk(u) = k(\|u\|)$ is a radial basis function with profile $k(z) = \exp \left(\max\{0, z - 1\}^2/4\right)$, chosen so that neighbour disks are centrally flat and then decay smoothly.

\subsection{Qualitative comparison}

\input{fig-compare-with-perturb}

\Cref{fig:comparison} shows a qualitative comparison between our method and others.
We see that our criterion of $\Phi_0 = \Phi(x)$ chooses fairly well-localized masks in most cases.
Masks tend to cover objects tightly, are sharp, and clearly identify a region of interest in the image.
\Cref{fig:area_progressions} shows what the network considered to be most discriminative ($a=5\%$; e.g., banjo fret board, elephant tusk) and complete ($a=20\%$) as the area increases.
We notice that evidence from several objects accumulates monotonically (e.g., impala and spider) and that foreground (e.g., ostrich) or discriminative parts (e.g., dog's nose) are usually sufficient.

In~\cref{fig:compare_with_perturb}, we compare our masks to those of Fong and Vedaldi~\cite{fong17interpretable}.
The stability offered by controlling the area of the perturbation is obvious in these examples.
Lastly, we visualize a sanity check proposed in Adebayo et al.~\cite{Adebayo2018Sanity} in~\cref{fig:sanity_check} (we use the ``hybrid'' formulation).
Unlike other backprop-based methods, our visualizations become significantly different upon weight randomization (see supp.~mat.~for more qualitative examples).

\input{fig-sanity-check}

\subsection{Pointing game}

\input{tbl-pointing-game}

A common approach to evaluate attribution methods is to correlate their output with semantic annotations in images.
Here we consider in particular the pointing game of Zhang et al.~\cite{zhang2016excitation}.
For this, an attribution method is used to compute a saliency map for each of the object classes present in the image.
One scores a hit if the maximum point in the saliency map is contained within the object;
The overall accuracy is the number of hits over number of hits plus misses.

\Cref{tbl:pointing_game} shows results for this metric and compares our method against the most relevant work in the literature on PASCAL VOC~\cite{Everingham15} (using the 2007 test set of $4952$ images) and COCO~\cite{lin2014microsoft} (using the 2014 validation set of $\approx 50k$ images).
We see that our method is competitive with VGG16 and ResNet50 networks.
In contrast, Fong and Vedaldi's~\cite{fong17interpretable} was not competitive in this benchmark (although they reported results using GoogLeNet).

\paragraph{Implementation details.}

Since our masks are binary, there is no well defined maximum point.
To apply our method to the pointing game, we thus run it for areas $\{0.025, 0.05, 0.1, 0.2\}$ for PASCAL and $\{0.018, 0.025, 0.05, 0.1\}$ for COCO (due to the smaller objects in this dataset).
The binary masks are summed and a Gaussian filter with standard deviation equal to 9\% of the shorter side of the image applied to the result to convert it to a saliency map.
We use  the original Caffe models of~\cite{zhang2016excitation} converted to PyTorch and use the preservation formulation of our method.

\subsection{Monotonicity of visual evidence}

\Cref{e:areac} implements the ``preservation game'' and searches for regions of a given area that \emph{maximally activate} the networks' output.
When this output is the confidence score for a class, we hypothesise that hiding evidence from the network would only make the confidence lower, i.e., we would expect the effect of maximal perturbations to be ordered consistently with their size:
\begin{equation}\label{e:monotonicity}
a_1 \leq a_2
~\Rightarrow~
\Phi(\bbm_{a_1} \otimes \bx) \leq \Phi(\bbm_{a_2} \otimes \bx)
\end{equation}
However, this may not always be the case.
In order to quantify the frequency of this effect, we test whether~\cref{e:monotonicity} holds for all $a_1, a_2 < a^*$, where $a^*$ is the optimal area of the extremal perturbation (\cref{e:areamax}, where $\Phi_0 = \Phi(\bx)$).
Empirically, we found that this holds for 98.45\% of ImageNet validation images, which indicates that evidence is in most cases integrated monotonically by the network.

More generally, our perturbations allow us to sort and investigate how information is integrated by the model in order of importance.
This is shown in several examples in~\cref{fig:area_progressions} where, as the area of the mask is progressively increased, different parts of the objects are prioritised.

%% file: fig-compare-with-perturb.tex
\begin{figure}
    \centering
    \includegraphics[width=0.24\linewidth, trim=5 5 5 5]{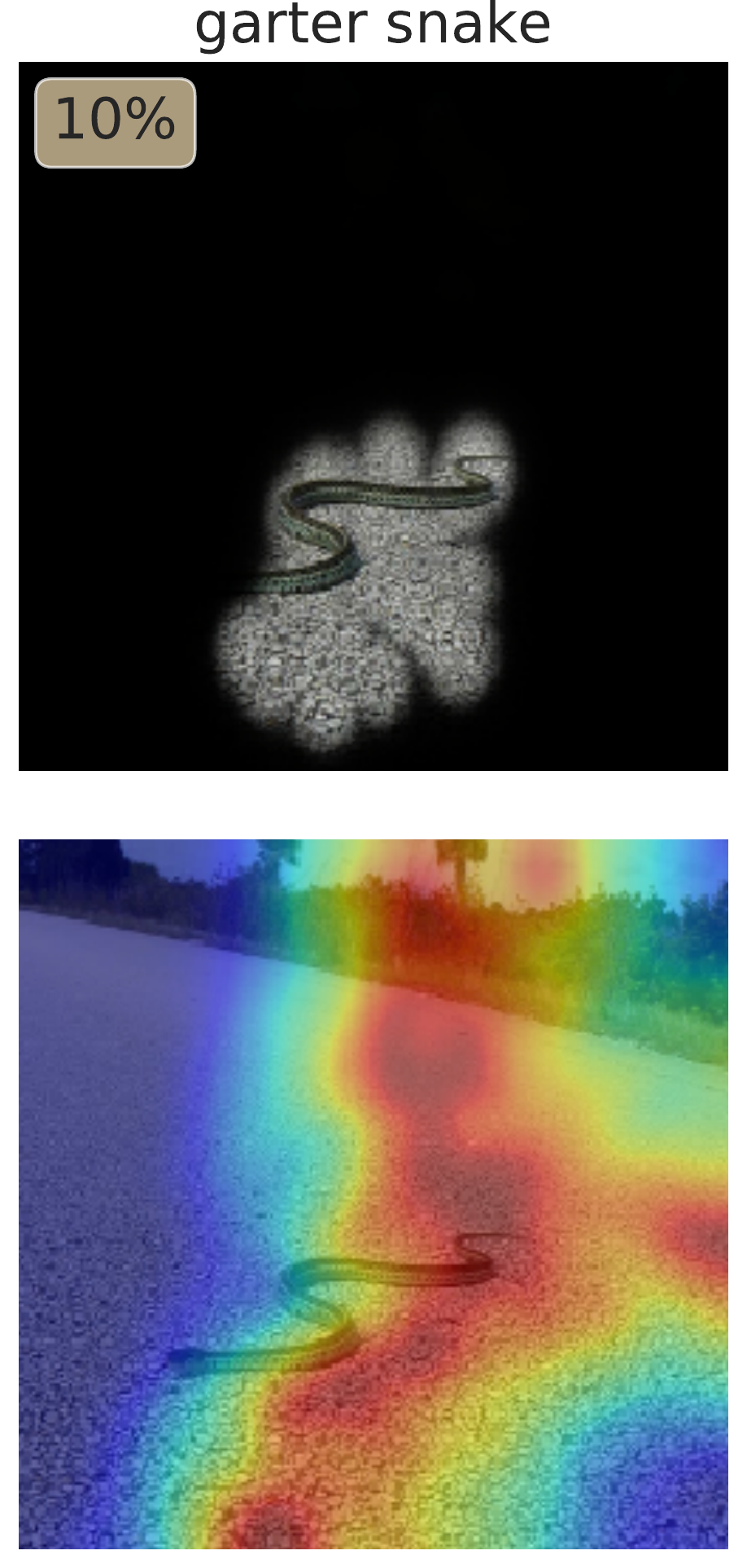}
    \includegraphics[width=0.24\linewidth, trim=5 5 5 5]{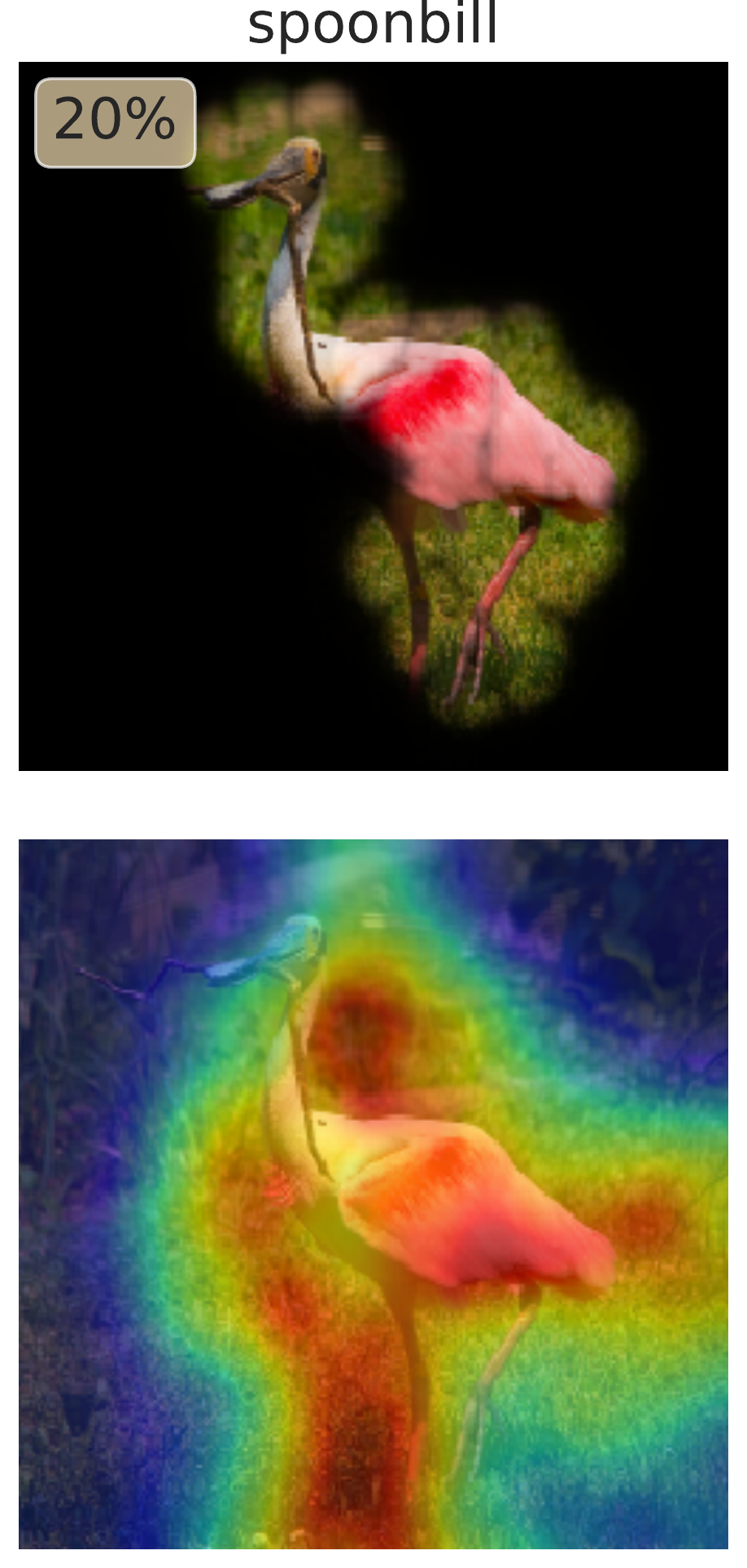}
    \includegraphics[width=0.24\linewidth, trim=5 5 5 5]{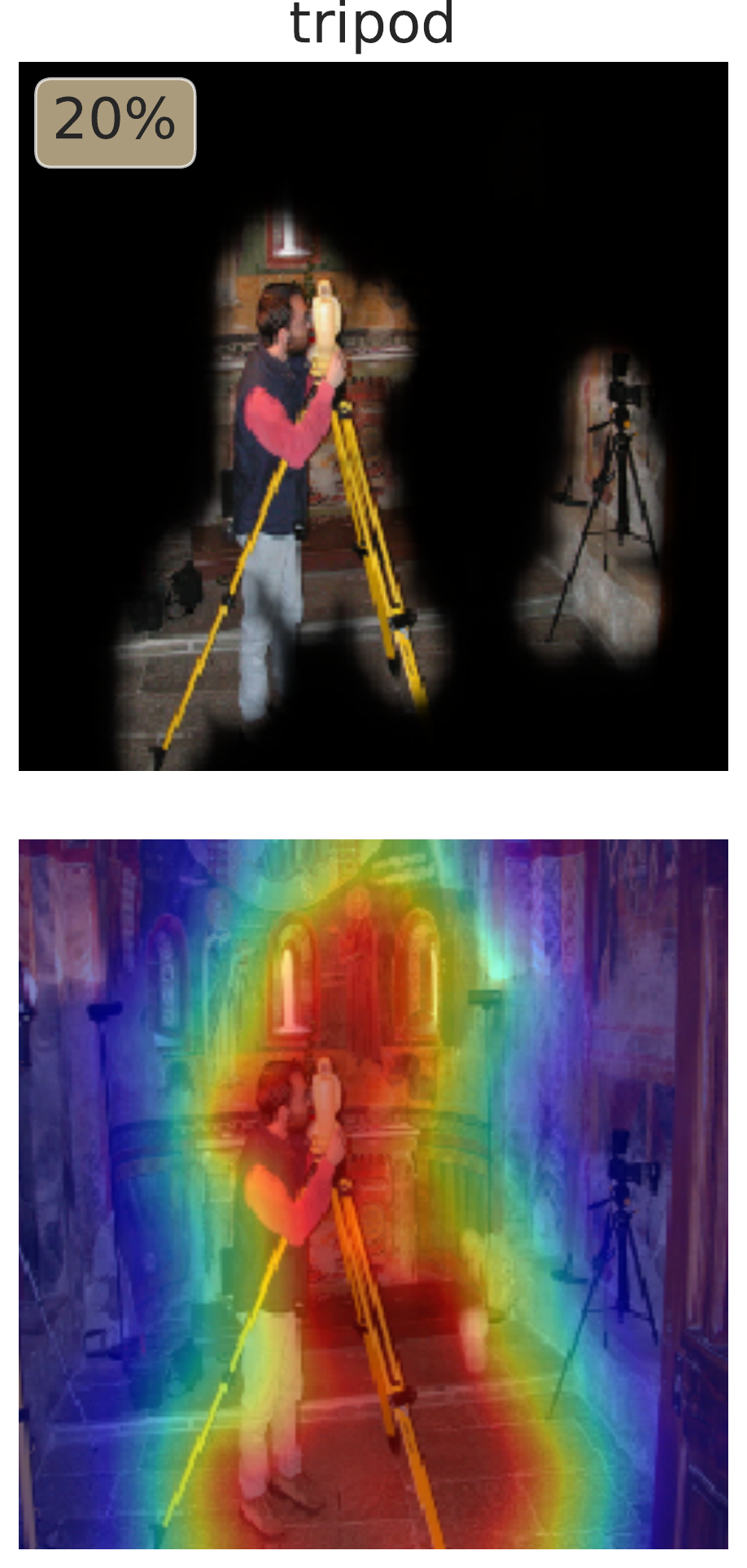}
    \includegraphics[width=0.24\linewidth, trim=5 5 5 5]{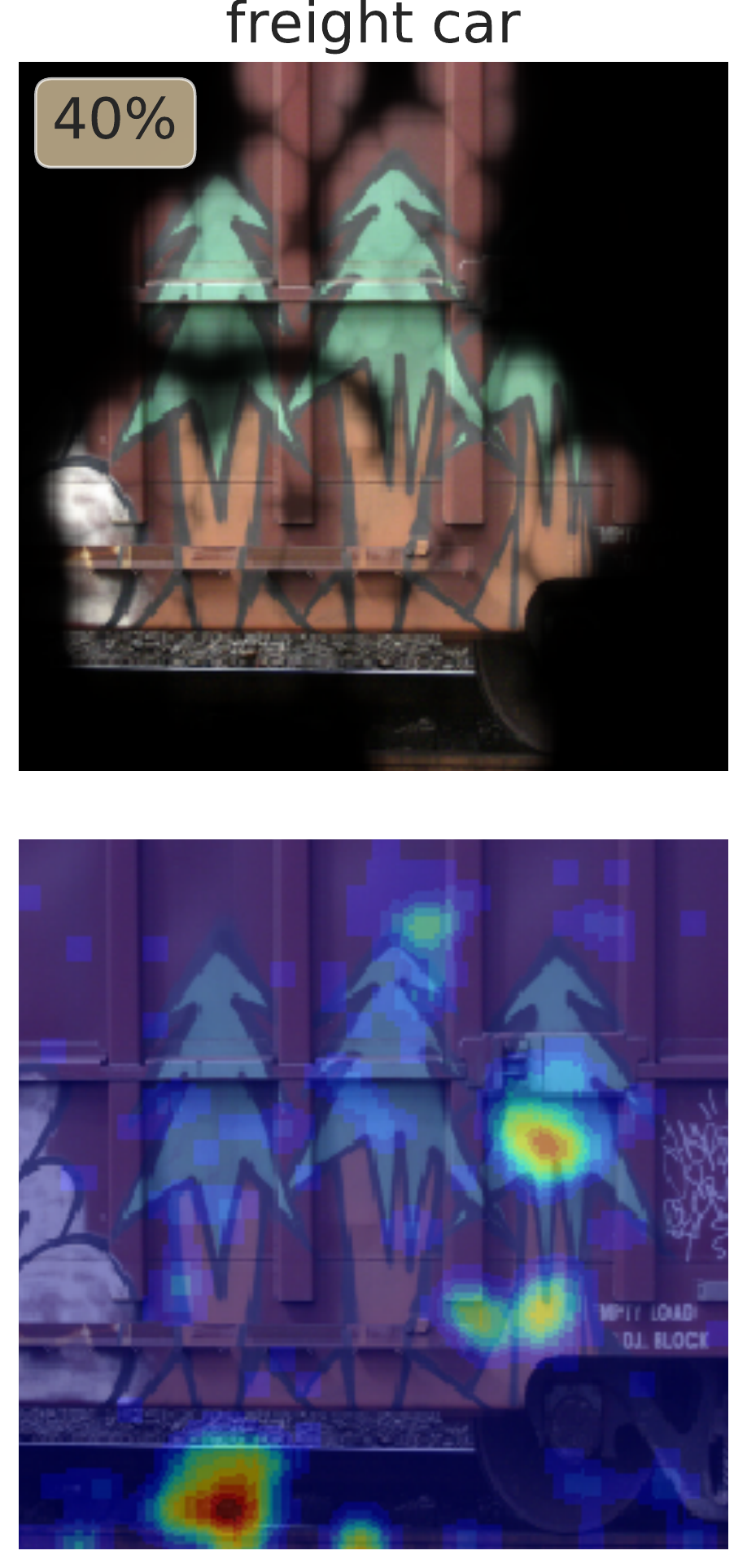}
    \caption{\textbf{Comparison with~\cite{fong17interpretable}.} Our extremal perturbations (top) vs. masks from Fong and Vedaldi~\cite{fong17interpretable} (bottom).}\label{fig:compare_with_perturb}
\end{figure}

%% file: fig-sanity-check.tex
\begin{figure}
    \centering
    \includegraphics[width=\linewidth]{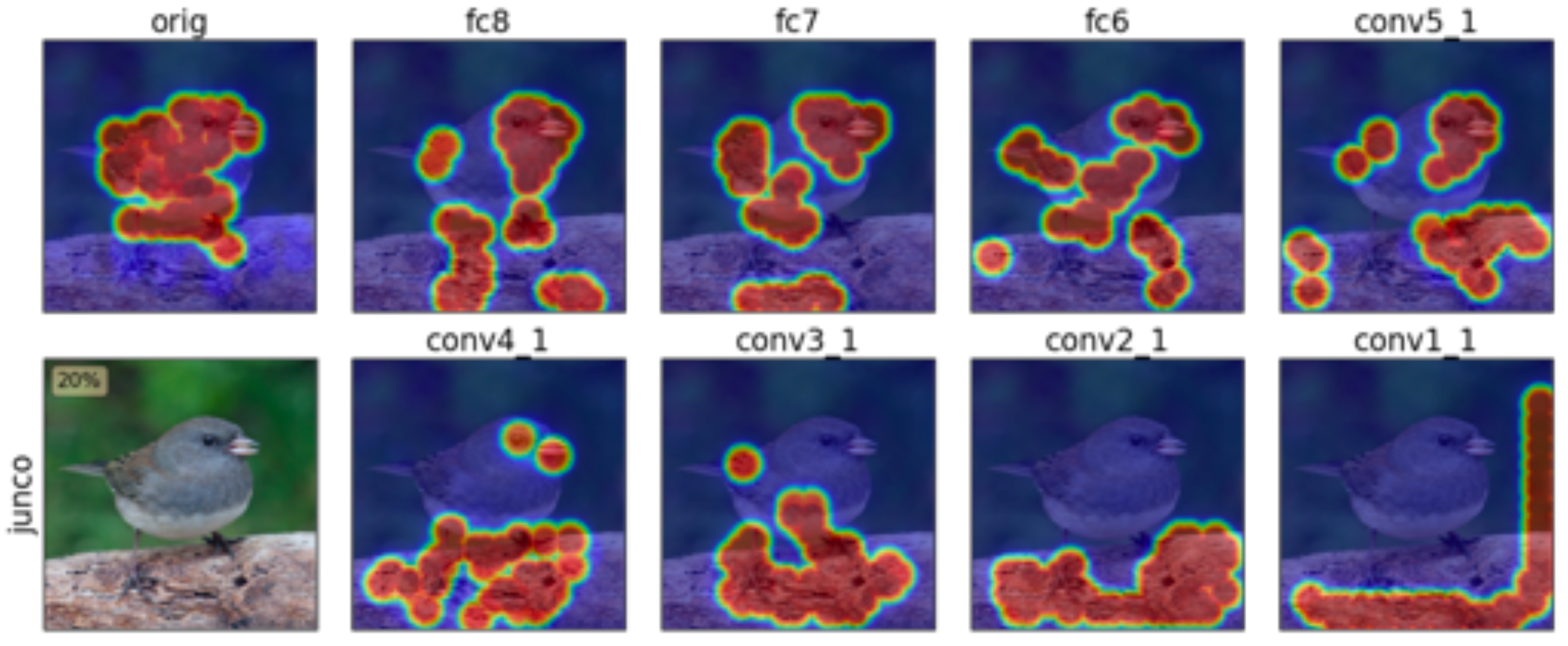}
    \caption{\textbf{Sanity check~\cite{Adebayo2018Sanity}.} Model weights are progressively randomized from fc8 to conv1\_1 in VGG16, demonstrating our method's sensitivity to model weights.}
    \label{fig:sanity_check}
\end{figure}

%% file: tbl-pointing-game.tex
\begin{table}
\centering
\setlength{\tabcolsep}{2pt}
\begin{tabular}{lcccc}
\toprule
       & \multicolumn{2}{c}{\emph{VOC07 Test (All/Diff)}}       & \multicolumn{2}{c}{\emph{COCO14 Val (All/Diff)}} \\
\cmidrule(lr){2-3} \cmidrule(lr){4-5}
Method & \textit{VGG16}                                         & \textit{ResNet50}                 & \textit{VGG16}       & \textit{ResNet50}     \\
\midrule
Cntr. & 	69.6/42.4 &	69.6/42.4 &	27.8/19.5 &	27.8/19.5 \\
Grad  & 	76.3/56.9 &	72.3/56.8 &	37.7/31.4 &	35.0/29.4 \\
DConv & 	67.5/44.2 &	68.6/44.7 &	30.7/23.0 &	30.0/21.9 \\
Guid. & 	75.9/53.0 &	77.2/59.4 &	39.1/31.4 &	42.1/35.3 \\
MWP   & 	77.1/56.6 &	84.4/70.8 &	39.8/32.8 &	49.6/43.9 \\
cMWP  & 	79.9/66.5 &\b{90.7}/\u{82.1}&49.7/44.3&\b{58.5}/\b{53.6}\\
RISE*& \u{86.9}/\u{75.1} & 86.4/78.8 & 50.8/45.3 & 54.7/50.0\\
GCAM&86.6/74.0&\u{90.4}/\b{82.3}&\b{54.2}/\b{49.0}&\u{57.3}/\u{52.3}\\
Ours* & \b{88.0}/\b{76.1} & 88.9/78.7 & \u{51.5}/\u{45.9} & 56.5/51.5 \\

\bottomrule
\end{tabular}
\caption{\textbf{Pointing game.} Mean accuracy on the pointing game over the full data splits and a subset of difficult images (defined in~\cite{zhang2016excitation}).
Results from PyTorch re-implementation using TorchRay package (* denotes average over 3 runs).
}\label{tbl:pointing_game}
\end{table}

%% file: intermediate.tex
\section{Attribution at intermediate layers}\label{s:intermediate}

\input{fig-area-curve-intermediate}

Lastly, we extend extremal perturbations to the \emph{direct} study of the intermediate layers in neural networks. This allows us to highlight a novel use case of our area loss and introduce a new technique for understanding which channels are salient for classification.

As an illustration, we consider in particular channel-wise perturbations.
Let $\Phi_{l}(x) \in \mathbb{R}^{K_{l}\times H_{l} \times W_{l} }$ be the intermediate representation computed by a neural network $\Phi$ up to layer $l$ and let $\Phi_{l+} : \mathbb{R}^{K_{l}\times H_{l} \times W_{l}} \to \mathbb{R}$ represent the rest of model, from layer $l$ to the last layer.
We then re-formulate the preservation game from~\cref{e:areac-relaxed} as:
\begin{equation}\label{e:intermediate_loss}
\bbm_a = \operatornamewithlimits{argmax}_{\bbm} \Phi_{l+}(\bbm \otimes \Phi_{l}(\bx)) - \lambda R_a(\bbm).
\end{equation}
Here, the mask $\bbm \in [0,1]^{K_l}$ is a vector with one element per channel which element-wise multiplies with the activations $\Phi_{l}(\bx)$, broadcasting values along the spatial dimensions.
Then, the extremal perturbation $\bbm_{a^*}$ is selected by choosing the optimal area
\begin{equation}\label{e:optimal_area_deletion}
    a^* = \min \{a: \Phi_{l+}(\bbm_a \otimes \Phi_{l}(\bx)) \geq \Phi_0  \}.
\end{equation}
We assume that the output $\Phi_{l+}$ is the pre-softmax score for a certain image class and we set the $\Phi_0 = \Phi(\bx)$ to be the model's predicted value on the unperturbed input (\cref{fig:area_curve_intermediate}).

\paragraph{Implementation details.}

In these experiments, we use GoogLeNet~\cite{szegedy2015going} and focus on layer $l =$\texttt{inception4d}, where $H_{l} = 14, W_l = 14, K_l = 528$.
We optimize \cref{e:intermediate_loss} for 300 iterations with a learning rate of $10^{-2}$.
The parameter $\lambda$ linearly increases from $0$ to $1500$ during the first 150 iterations, after which $\lambda = 1500$ stays constant.
We generate channel-wise perturbation masks for areas $a\in\{1, 5, 10, 20, 25, 35, 40, 50, 60, 70, 80, 90, 100, 150, 200, \\250, 300, 350, 400, 450, 528\}$, where $a$ denotes the number of channels preserved.

The saliency heatmaps in~\cref{fig:area_curve_intermediate} and~\cref{fig:intermediate_feature_inversions} for channel-wise attribution are generated by summing over the channel dimension the element-wise product of the channel attribution mask and activation tensor at layer $l$:
\begin{equation}\label{e:channel_overlay}
    \bm v = \sum_{k \in K} \bbm_{a^*}^k \otimes \Phi_{l}^k(\bx)
\end{equation}

\input{fig-intermediate-channel-inversions}

\subsection{Visualizing per-instance channel attribution}\label{s:per_instance_channel}

Unlike per-instance input-level spatial attribution, which can be visualized using a heatmap, per-instance intermediate channel attribution is more difficult to visualize because simply identifying important channels is not necessarily human-interpretable. To address this problem, we use feature inversion~\cite{mahendran16visualizing, olah2017feature} to find an image that maximises the dot product of the channel attribution vector and the activation tensor (see~\cite{olah2017feature} for more details):

\begin{equation}\label{e:feature_inversion}
    I^* = \argmax_I \{(\bbm_{a^*} \otimes \Phi_{l}(\bx)) \cdot \Phi_{l} (I)\}
\end{equation}
where $\bbm_a^*$ is optimal channel attribution mask at layer $l$ for input image $\bx$ and $\Phi_{l}(I)$ is the activation tensor at layer $l$ for image $I$, the image we are learning.

This inverted image allows us to identify the parts of the input image that are salient for a particular image to be correctly classified by a model.
We can compare the feature inversions of activation tensors perturbed with channel mask (right column in~\cref{fig:intermediate_feature_inversions}) to the inversions of original, unperturbed activation tensors (middle column) to get a clear idea of the most discriminative features of an image.

Since the masks are roughly binary, multiplying $\bbm_a^*$ with the activation tensor $\Phi_l(\bx)$ in~\cref{e:feature_inversion} zeroes out non-salient channels.
Thus, the differences in the feature inversions of original and perturbed activations in~\cref{fig:intermediate_feature_inversions} highlight regions encoded by salient channels identified in our attribution masks (i.e., the channels that are not zeroed out in~\cref{e:feature_inversion}).

\input{fig-intermediate-inversions}

\subsection{Visualizing per-class channel attribution}\label{s:per_class_channel}

We can also use channel attribution to identify important, class-specific channels.
In contrast to other methods, which explicitly aim to find class-specific channels and/or directions at a global level~\cite{fong18net2vec, kim2017interpretability, zhou2018revisiting}, we are able to similarly do so ``for free'' using only our per-instance channel attribution masks.
After estimating an optimal masks $\bbm_{a^*}$ for all ImageNet validation images, we then create a per-class attribution mask $\bar{\bbm}_c \in [0,1]^K$ by averaging the optimal masks of all images in a given class $c$.
Then, we can identify the most important channel for a given class as follows: $k^*_c = \argmax_{k\in K} \bar{\bbm}_c^k$.
In~\cref{fig:intermediate_feature_inversions_channels}, we visualize two such  channels via feature inversions.
Qualitatively, these feature inversions of channels $k^*_c$ are highly class-specific.

%% file: fig-area-curve-intermediate.tex
\begin{figure}
\centering
\includegraphics[height=100pt]{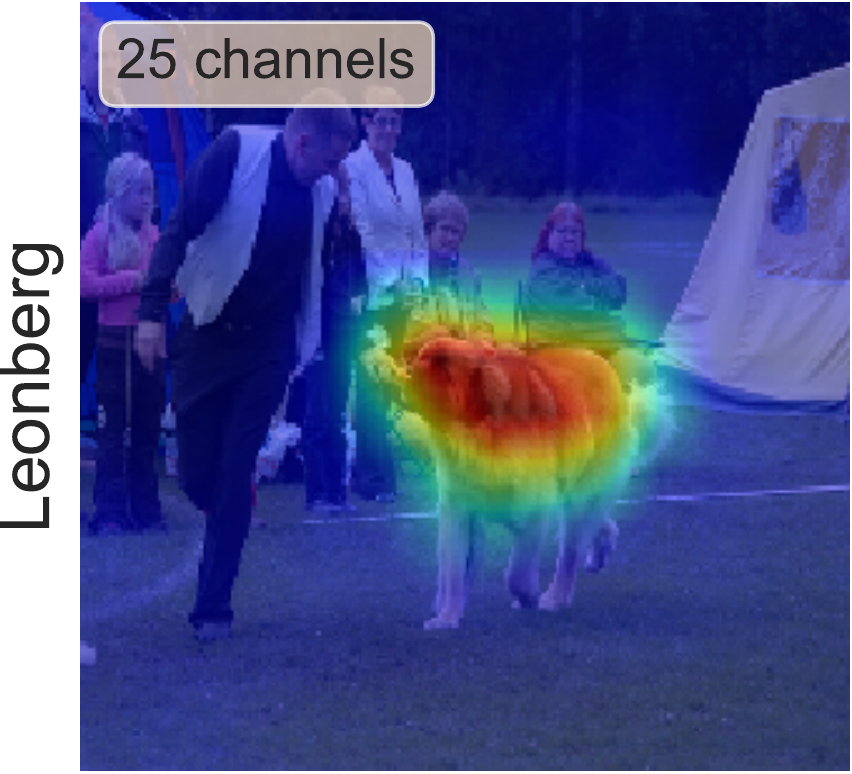}
\includegraphics[height=100pt]{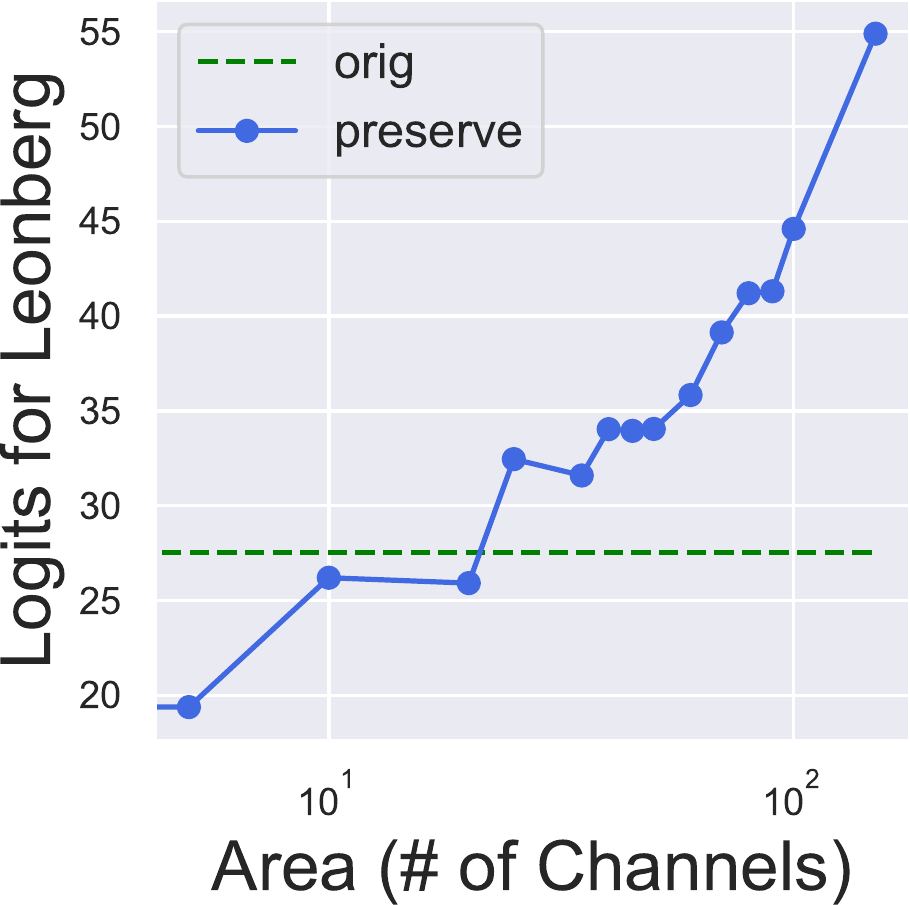}
\caption{\textbf{Attribution at intermediate layers.} Left: This is visualization (\cref{e:channel_overlay}) of the optimal channel attribution mask $\bbm_{a^*}$, where $a^* = 25$ channels, as defined in~\cref{e:optimal_area_deletion}. Right: This plot shows that class score monotonically increases as the area (as the number of channels) increases.
}\label{fig:area_curve_intermediate}
\end{figure}

%% file: fig-intermediate-channel-inversions.tex
\begin{figure}
\centering
\includegraphics[height=75pt]{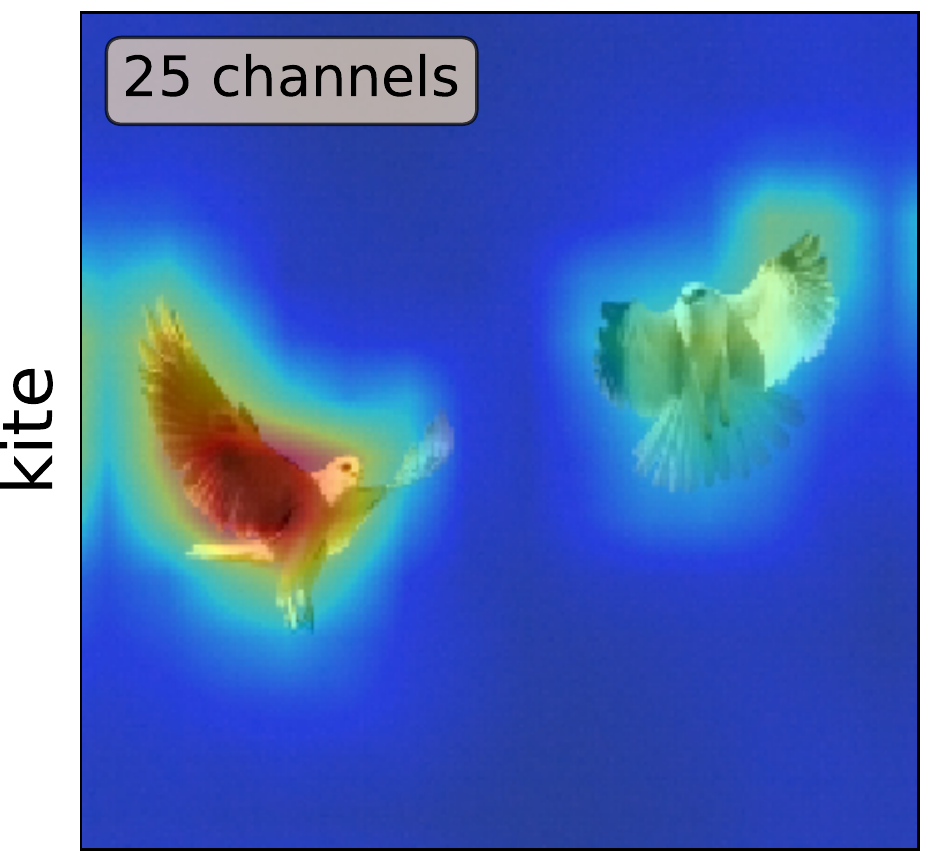}
\includegraphics[height=75pt]{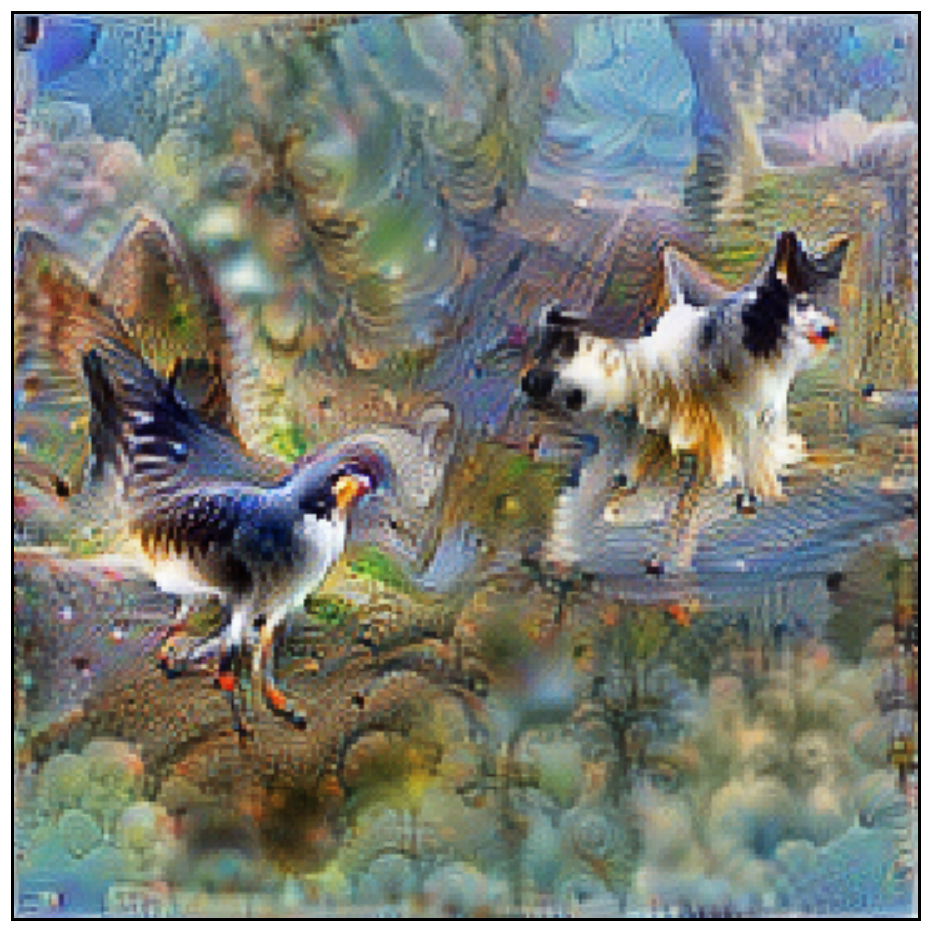}
\includegraphics[height=75pt]{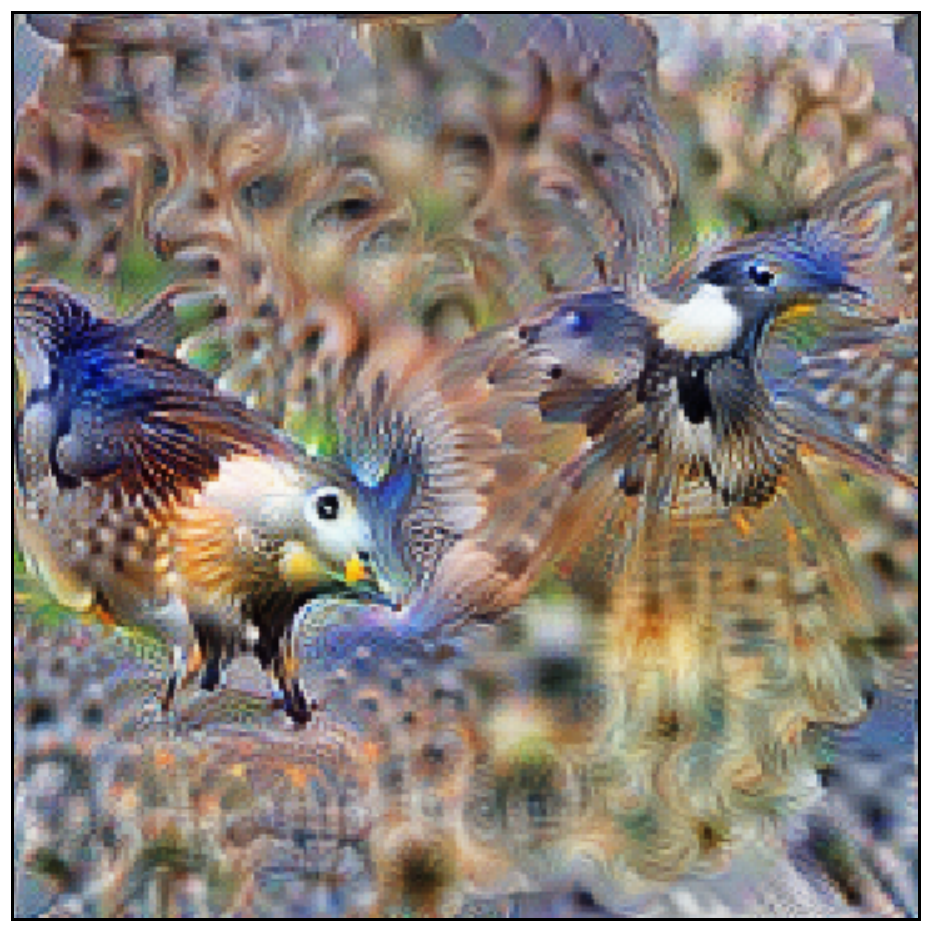}
\includegraphics[height=75pt]{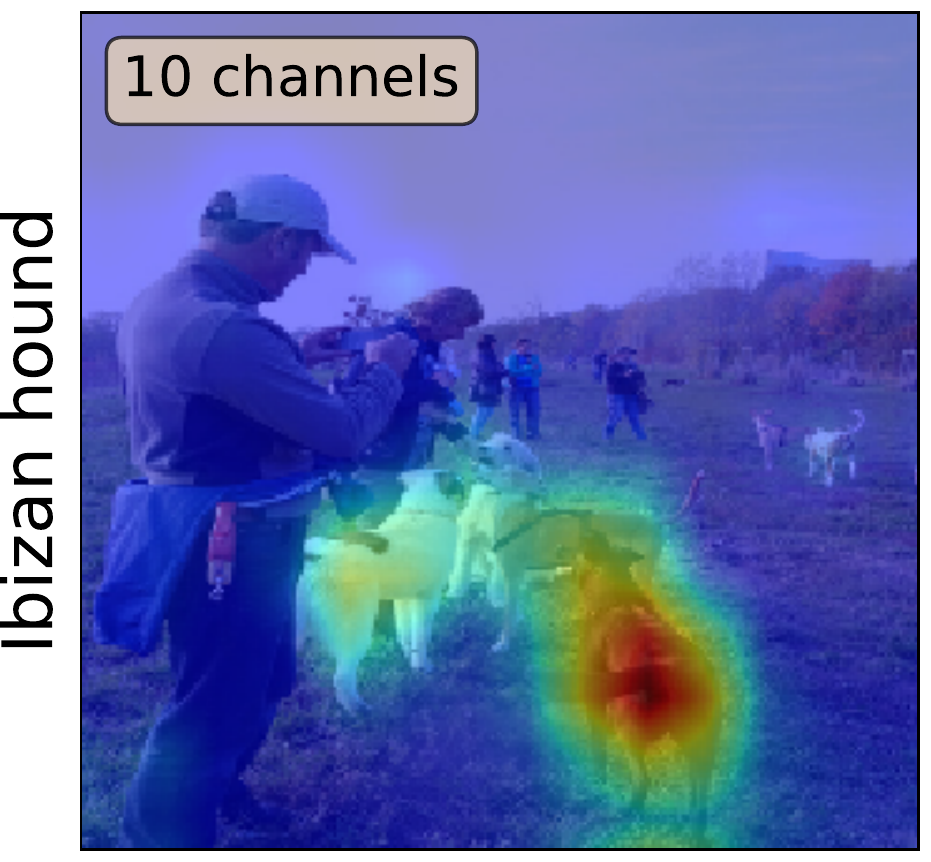}
\includegraphics[height=75pt]{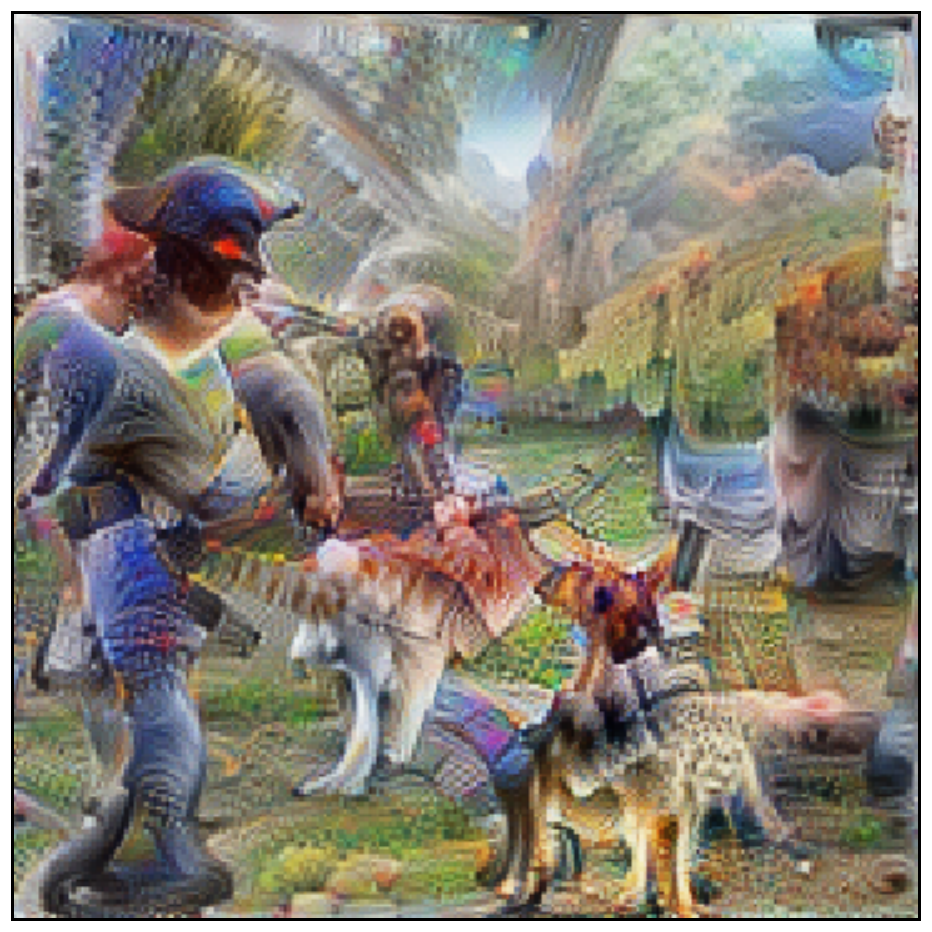}
\includegraphics[height=75pt]{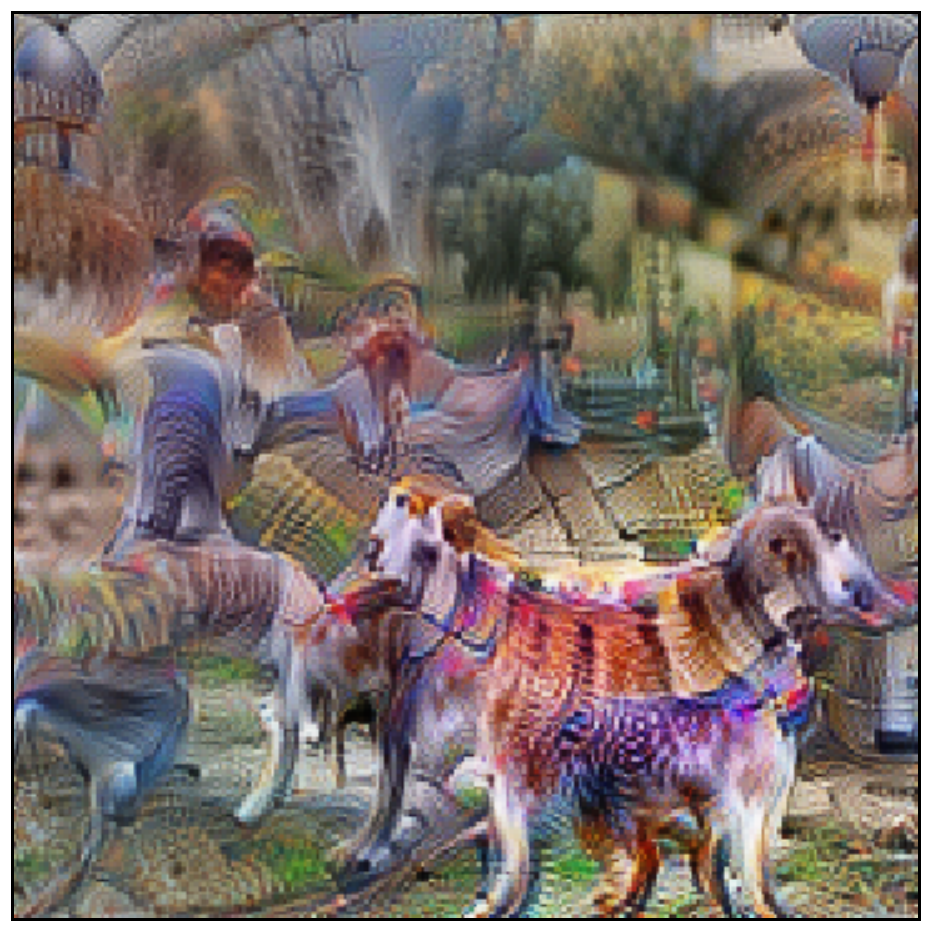}
\includegraphics[height=75pt]{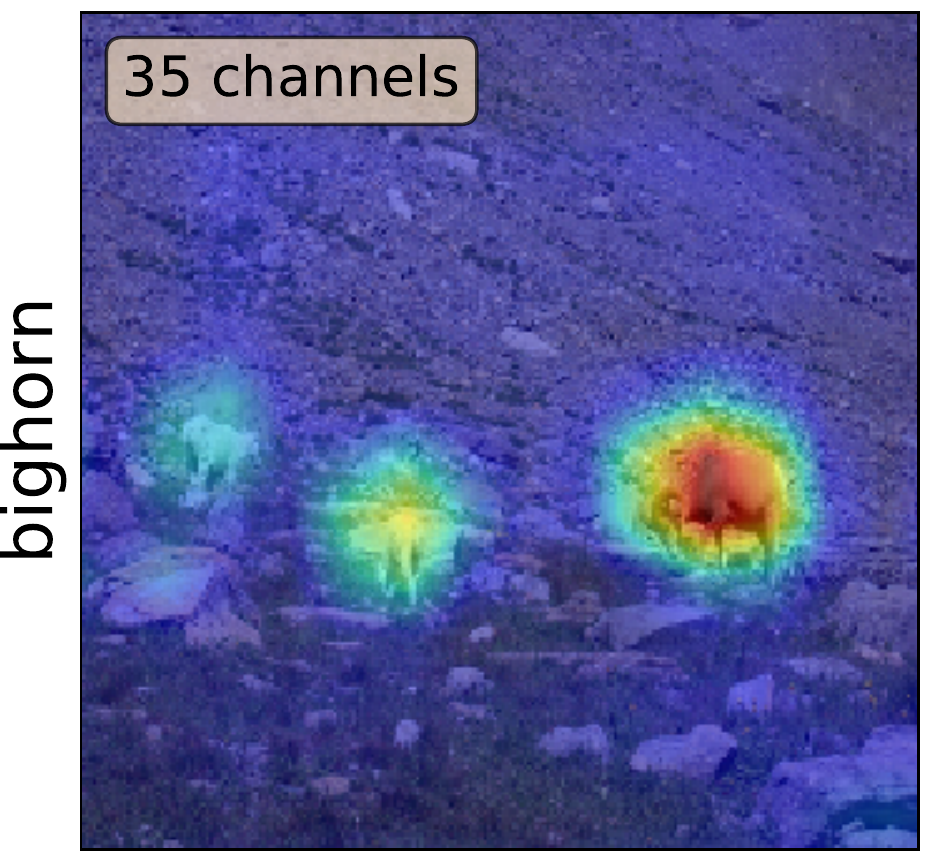}
\includegraphics[height=75pt]{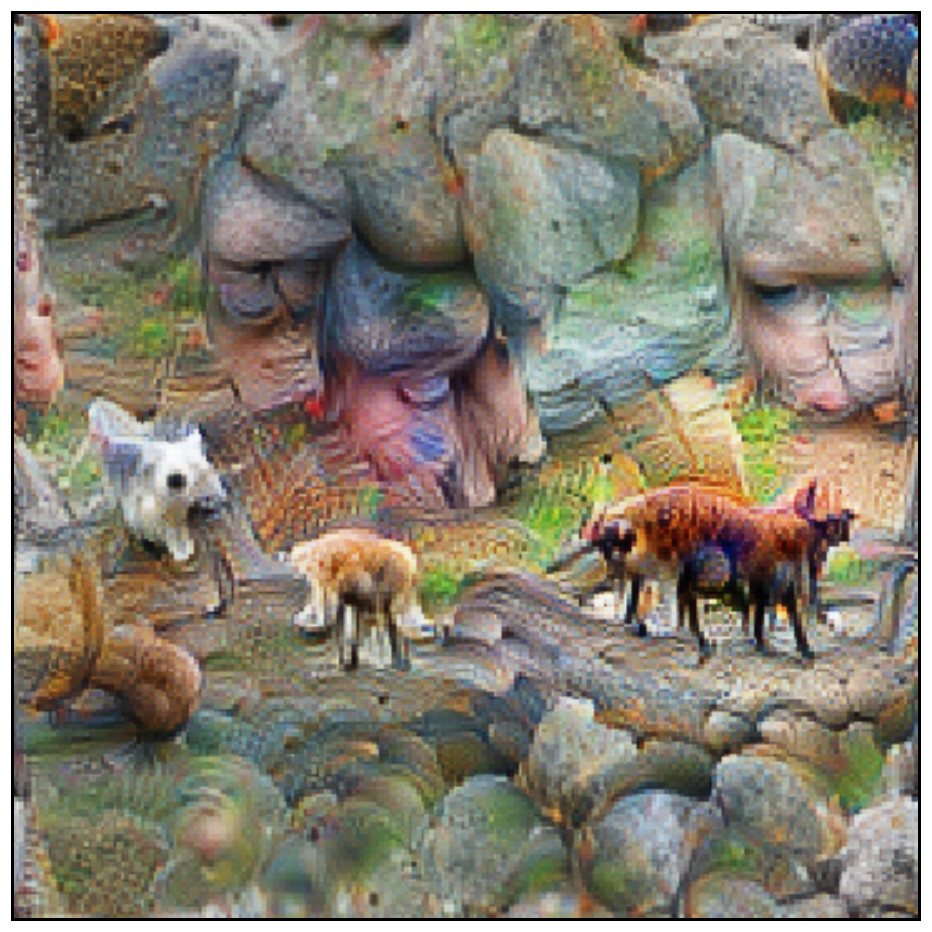}
\includegraphics[height=75pt]{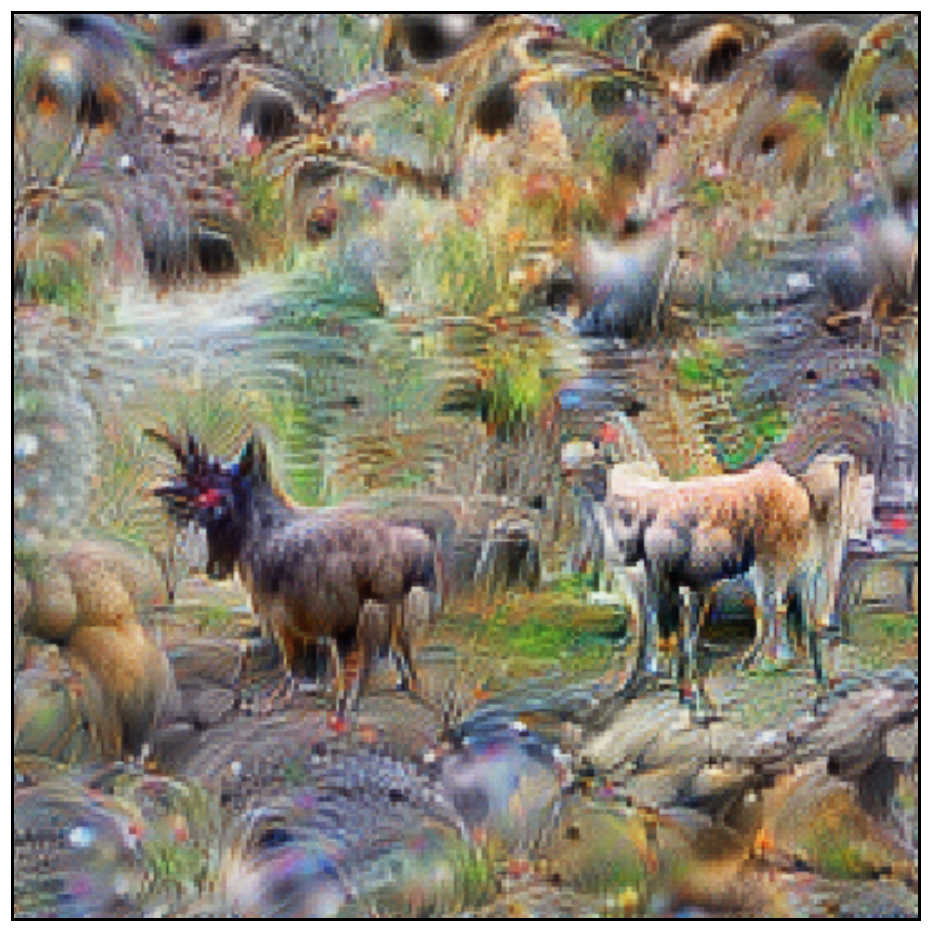}
\includegraphics[height=75pt]{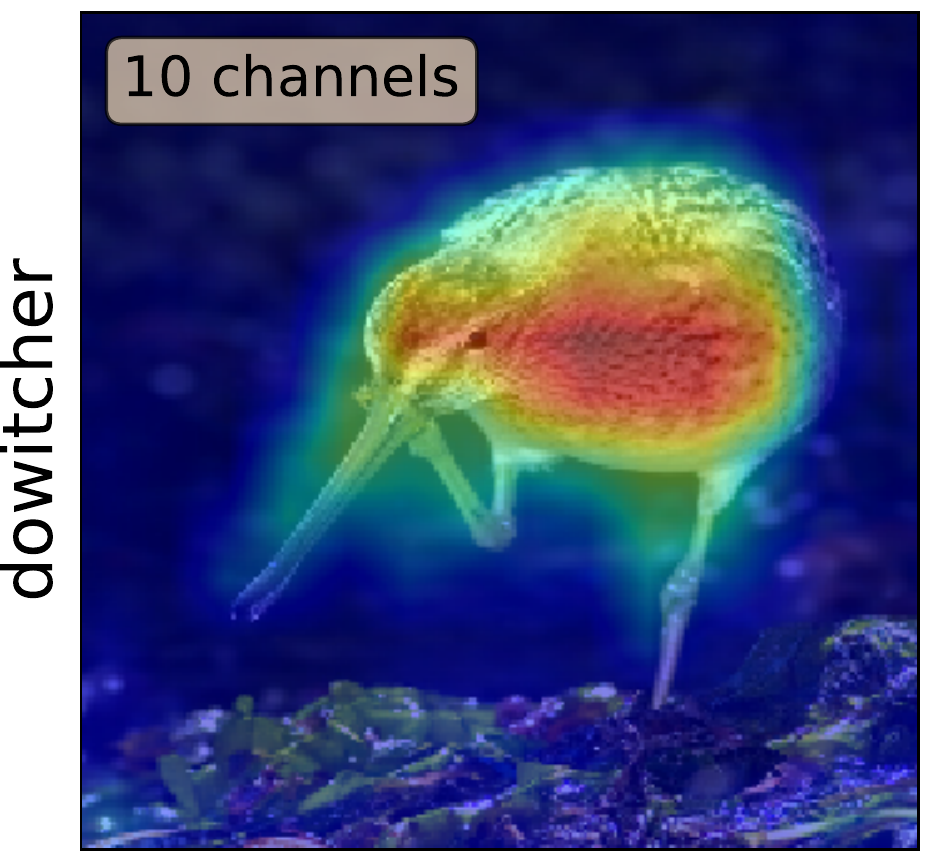}
\includegraphics[height=75pt]{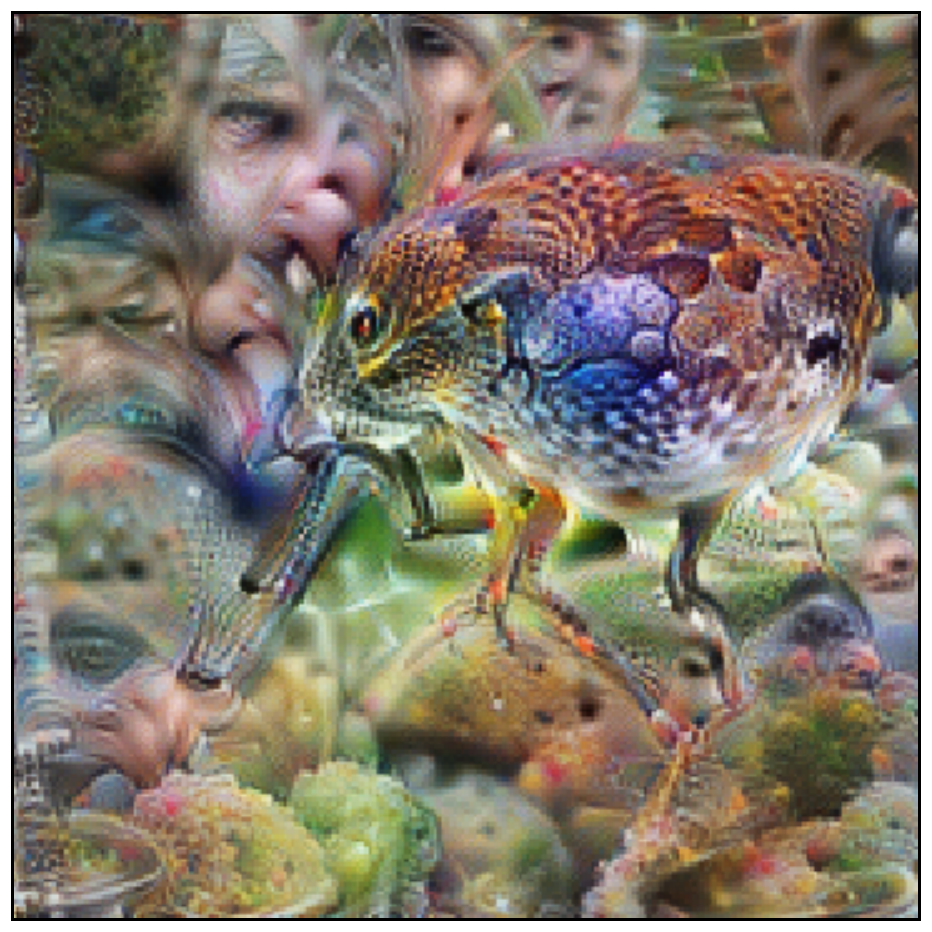}
\includegraphics[height=75pt]{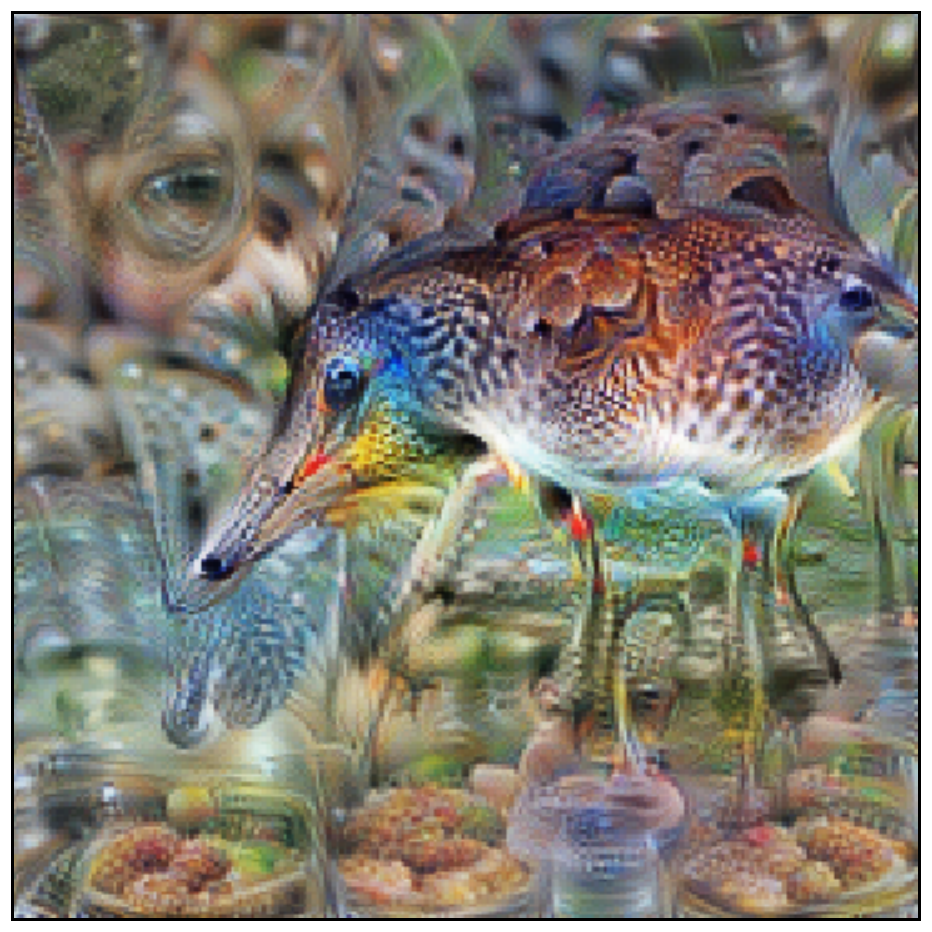}
\caption{\textbf{Per-instance channel attribution visualization.} Left: input image overlaid with channel saliency map (\cref{e:channel_overlay}). Middle: feature inversion of original activation tensor. Right: feature inversion of activation tensor perturbed by optimal channel mask $\bbm_{a^*}$. By comparing the difference in feature inversions between un-perturbed (middle) and perturbed activations (right), we can identify the salient features that our method highlights.}
\label{fig:intermediate_feature_inversions}
\end{figure}

%% file: fig-intermediate-inversions.tex
\begin{figure}
\centering
\includegraphics[width=\linewidth]{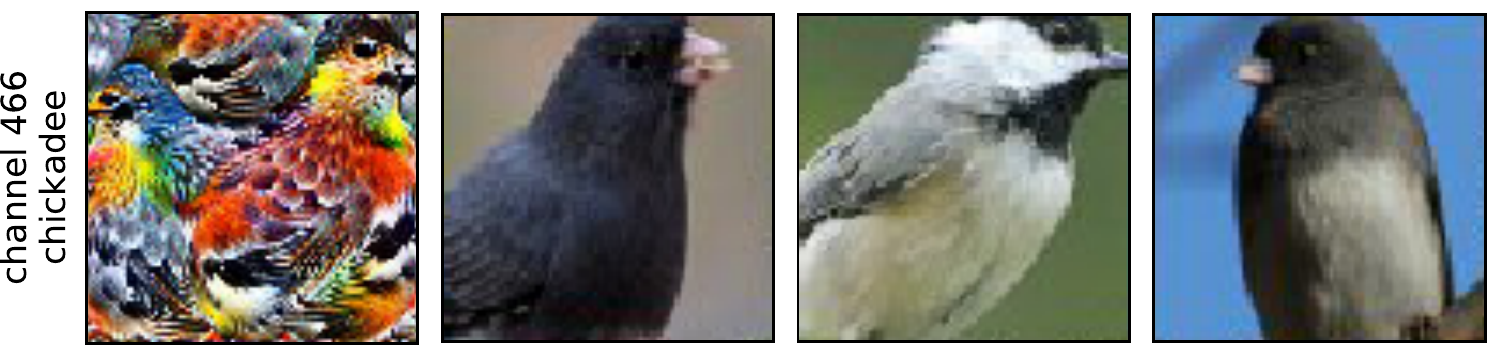}
\includegraphics[width=\linewidth]{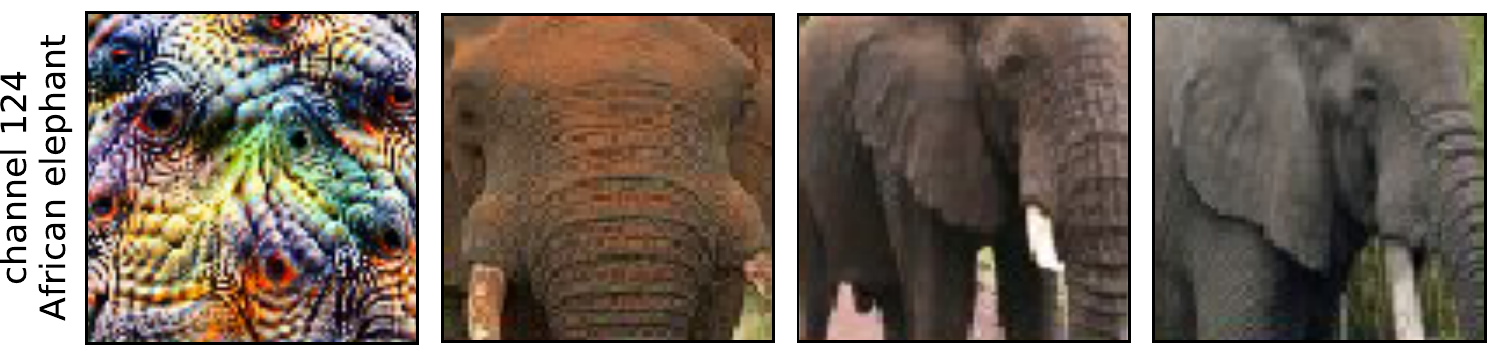}
\caption{\textbf{Discovery of salient, class-specific channels.} By analyzing $\bar{\bbm}_c$, the average over all $\bbm_{a^*}$ for class $c$ (see~\cref{s:per_class_channel}), we automatically find salient, class-specific channels like these. First column: channel feature inversions; all others: dataset examples.}
\label{fig:intermediate_feature_inversions_channels}
\end{figure}

%% file: conclusions.tex
\section{Conclusion}\label{s:conclusions}
We have introduced the framework of extremal perturbation analysis, which avoids some of the issues of prior work that use perturbations to analyse neural networks.
We have also presented a few technical contributions to compute such extremal perturbation.
Among those, the rank-order area constraint can have several other applications in machine learning beyond the computation of extremal perturbations.
We have extended the perturbations framework to perturbing intermediate activations and used this to explore a number of properties of the representation captured by a model.
In particular, we have visualized, likely for the first time, the difference between perturbed and unperturbed activations using a representation inversion technique.
Lastly, we released TorchRay~\cite{torchray19}, a PyTorch interpretability library in which we've re-implemented popular methods and benchmarks to encourage reproducible research.

%% file: appendix.tex
\appendix
\section{Implementation details}\label{s:details}

\subsection{Generating smooth masks}\label{s:details.generator}

We implement the equation:
$$
  \hat m(u) = \operatorname{pool}_i g(u - u_i) m_i
$$
Here $i=0,\dots,N-1$ are samples of the input mask, $u=0,\dots,W-1$ samples ouf the output mask, and $u_i$ is the mapping between input and output samples, $u_i = a i + b$.
We assume that the kernel $k$ has a ``radius'' $\sigma$, in the sense that only samples $|u - u_i|\leq \sigma$ matter.

In order to compute this expansion fast, we use the unpool operator.
In order to do so, unpool is applied to $m$ with window $K=2R+1$ and padding $P$.
This results in the signal
\begin{multline*}
    m'_{k,i} = m_{i+k-P},
    ~~~0\leq i \leq W'-1,
    ~~~0\leq k \leq K-1,\\
    W'=N-K+2P+1.
\end{multline*}
We then use nearest-neighbour upsampling in order to bring this signal in line with the resolution of the output:
\begin{multline*}
    m''_{k,u} = m_{\lfloor \frac{u}{s} \rfloor+k-P},
    ~~~~
    0\leq u \leq W''-1, \\
    0\leq k \leq K-1.
\end{multline*}
Here the upsampling factor is given by $s = W''/W'$.
In PyTorch, we specify upsampling via the input size $W'$ and the output size $W''$, so we need to choose $W''$ appropriately.

To conclude, we do so as follows.
We choose a $\sigma$ (kernel width in image pixels) and and $s$ (kernel step in pixels).
We also choose a margin $b \geq 0$ to avoid border effects and set $a=s$.
With this, we see that computing $\hat m(u)$ requires samples:
\begin{multline*}
    u - \sigma \leq u_i \leq u + \sigma\\
    \Leftrightarrow~~~
    \frac{u}{s} - \frac{\sigma+b}{s} \leq i \leq \frac{u}{s} + \frac{\sigma-b}{s}.
\end{multline*}
On the other hand, at location $u$ in $m''_{k,u}$ we have pooled samples $m_i$ for which:
$$
\left\lfloor \frac{u}{s} \right\rfloor-P \leq i \leq \left\lfloor \frac{u}{s} \right\rfloor+K-1-P.
$$
Hence we require
$$
\left\lfloor \frac{u}{s} \right\rfloor-P \leq \frac{u}{s} - \frac{\sigma+b}{s}
~~~\Rightarrow~~~
P \geq \frac{\sigma+b}{s} + \left\lfloor \frac{u}{s}\right\rfloor - \frac{u}{s}.
$$
Conservatively, we take:
$$
P = 1 + \left\lceil \frac{\sigma+b}{s}\right\rceil
$$
The other bound is:
$$
\frac{u}{s} + \frac{\sigma-b}{s} \leq \left\lfloor \frac{u}{s} \right\rfloor+K-1-P.
$$
Hence:
$$
K \geq \frac{u}{s} - \left\lfloor \frac{u}{s} \right\rfloor + \frac{\sigma-b}{s} + P + 1
$$
Hence, conservatively we take:
$$
K = 3 
+ \left\lceil \frac{\sigma+b}{s}\right\rceil
+ \left\lceil \frac{\sigma-b}{s}\right\rceil.
$$
Since $K=2R +1$ and $b\approx\sigma$, we set
$$
R = 1 + \left\lceil \frac{\sigma}{s}\right\rceil.
$$
In this way, we obtain a pooled mask:
$$
\bar m(u)
=
\operatornamewithlimits{pool}_{i}
g(u - u_i)~m_i
=
\operatornamewithlimits{pool}_{0 \leq k \leq K-1}
g_{k,u} m''_{k,u},
$$
where
$$
g_{k,u} = g(u - \bar u(u,k)),
~~~
\bar u(u,k)
=
\left\lfloor \frac{u}{s} \right\rfloor+k-P.
$$
Hence, the steps are: given the input mask parameters $m_i$, use unpooling to obtain $m'_{k,i}$ and then upsampling to obtain $m''_{k,u}$. Then use the equation above to pool using a pre-computed weights $g_{k,u}$.

Generally, the input to the mask generator are: $s$, $\sigma$ and the desired mask $\hat m(u)$ width $W$.
So far, we have obtained a mask $\bar m(u)$ with with $W''$, where $W'' = s W'$ is chosen to obtain the correct scaling factor and $W' = N - K + 2P +1$.
As a rule of thumb, we set $N = \lceil W / s \rceil$ in order to spread the $N$ samples at regular interval over the full stretch $W$.
We then set $R, K, P, W'$ and $W''$ according to the formulas above.
Once $\bar m(u)$ is obtained, we take a $W$-sized crop shifted by $b$ pixels to obtain the final mask $\hat m(u)$.

\section{Supplementary Materials}
The full supplementary materials for this paper can be found at \url{ruthcfong.github.io/files/fong19_extremal_supps.pdf}.